\documentclass[letterpaper,10pt,twocolumn]{article}
\usepackage[dvips]{graphicx}
\usepackage{epsfig}
\usepackage{theorem}
\usepackage{subeqn}
\newcommand{\gbf}[1] {\mbox{\boldmath${#1}$\unboldmath}}
\newcommand{\be}{\begin{equation}}
\newcommand{\ee}{\end{equation}}
\newcommand{\beq}{\begin{equation}}
\newcommand{\eeq}{\end{equation}}
\newcommand{\bed}{\begin{displaymath}}
\newcommand{\eed}{\end{displaymath}}
\newcommand{\beqa}{\begin{eqnarray}}
\newcommand{\eeqa}{\end{eqnarray}}
\newcommand{\beqann}{\begin{eqnarray*}}
\newcommand{\eeqann}{\end{eqnarray*}}
\newcommand{\bseq}{\begin{subequation}}
\newcommand{\eseq}{\end{subequation}}

\newcommand{\ba}{\begin{array}}
\newcommand{\ea}{\end{array}}

\newcommand{\negr}[1]{{\bf {#1}}}

\makeatletter
\theoremstyle{plain}
\newtheorem{Exa}{Example}
\newtheorem{Def}{Definition}
\newtheorem{Theo}{Theorem}
\newtheorem{Pro}{Proof}
\def\@normalsize{\@setsize\normalsize{12pt}\xpt\@xpt
\abovedisplayskip 10pt plus2pt minus5pt\belowdisplayskip
\abovedisplayskip
\abovedisplayshortskip \z@ plus3pt\belowdisplayshortskip 6pt plus3pt minus3pt
\let\@listi\@listI}
\def\subsize{\@setsize\subsize{12pt}\xipt\@xipt}
\def\section{\@startsection {section}{1}{\z@}{24pt plus 2pt minus 4pt}
{12pt plus 2pt minus 2pt}{\large\bf}}
\def\subsection{\@startsection {subsection}{2}{\z@}{12pt plus 2pt minus 4pt}
{12pt plus 2pt minus 2pt}{\subsize\bf}}
\makeatother
\font\tencmr=cmr10

\begin{document}
\setlength{\dblfloatsep}{10pt plus 1pt minus 5pt}
\setlength{\dbltextfloatsep}{10pt plus 1pt minus 5pt}
\date{}
\title {\Large\bf Working modes and aspects in fully parallel manipulators}
\author{\begin{tabular}[t]{c}
        Damien Chablat \hfill  Philippe Wenger \\
        {\tencmr Institut de Recherche en Cybern\'etique de Nantes} \\
        {\tencmr \'Ecole Centrale de Nantes} \\
        {\tencmr 1, rue de la No\"e, 44321 Nantes, France} \\
        {\tencmr Damien.Chablat@lan.ec-nantes.fr} \hfill
        {\tencmr Philippe.Wenger@lan.ec-nantes.fr} \\
\end{tabular}}
\maketitle
\thispagestyle{empty}
\thispagestyle{empty}
\thispagestyle{empty}
\subsection*{\centering Abstract}
{\em The aim of this paper is to characterize the notion of aspect in the workspace and in the joint space for parallel manipulators. In opposite to the serial manipulators, the parallel manipulators can admit not only multiple inverse kinematic solutions, but also multiple direct kinematic solutions. The notion of aspect introduced for serial manipulators in \cite{Borrel:86}, and redefined for parallel manipulators with only one inverse kinematic solution in \cite{Wenger:97}, is redefined for general fully parallel manipulators. Two Jacobian matrices appear in the kinematic relations between the joint-rate and the Cartesian-velocity vectors, which are called the ``inverse kinematics" and the ``direct kinematics" matrices. The study of these matrices allow to respectively define the parallel and the serial singularities. The notion of working modes is introduced to separate inverse kinematic solutions. Thus, we can find out domains of the workspace and the joint space exempt of singularity. Application of this study is the moveability analysis in the workspace of the manipulator as well as path-planing and control. This study is illustrated in this paper with a RR-RRR planar parallel manipulator.}
\begin{keyword}
Kinematics, Fully Parallel Manipulator, Aspects, Working modes,
Singularity.
\end{keyword}
\section{Introduction}
A well known feature of parallel manipulators is the existence of
multiple solutions to the direct kinematic problem. That is, the
mobile platform can admit several positions and orientations (or
configurations) in the workspace for one given set of input joint
values \cite{Merlet:97}. Moreover, parallel manipulators exist with
multiple inverse kinematic solutions. This means that the mobile
platform can admit several input joint values corresponding to one
given configuration of the end-effector. To cope with the existence
of multiple inverse kinematic solutions in {\em serial}
manipulators, the notion of aspects was introduced in
\cite{Borrel:86}. The aspects equal the maximal singularity-free
domains in the joint space. For usual industrial serial
manipulators, the aspects were found to be the maximal sets in the
joint space where there is only one inverse kinematic solution.
\par
A definition of the notion of aspect was given by \cite{Wenger:97}
for parallel manipulators with only one inverse kinematic solution.
These aspects were defined as the maximal singularity-free domains
in the workspace. For instance, this definition can apply to the
Stewart platform \cite{Stewart:65}.
\par
First of all, the working modes are introduced to allow the
separation of the inverse kinematic solutions. Then, a general
definition of the notion of aspect is given for all fully parallel
manipulators. The new aspects are the maximal singularity-free
domains of the Cartesian product of the workspace with the joint
space.
\par
A possible use of these aspects are the determination of the best
working mode. It allows to achieve complex task in the workspace or
to make path-planing without collision. As a matter of fact,
currently, the parallel manipulators possessing multiple inverse
kinematic solutions evolve only in one working mode. For a given
working mode, the aspect associated is different. It is possible to
choose one or several working modes to execute the tasks expected
in the maximal workspace of the manipulator.
\section{Preliminaries}
In this paragraph, some definitions permitting to introduce the
general notion of aspect are quoted.
\subsection{The fully parallel manipulators}
\begin{Def}
A fully parallel manipulator is a mechanism that includes as many
elementary kinematic chains as the mobile platform does admit
degrees of freedom. Moreover, every elementary kinematic chain
possesses only one actuated joint (prismatic, pivot or kneecap).
Besides, no segment of an elementary kinematic chain can be linked
to more than two bodies \cite{Merlet:97}.
\label{Definition:Fully_Parallel_Manipulator}
\end{Def}
In this study, kinematic chains are always independent. This
condition is necessary to find the working modes. Also, the
elementary kinematic chains can be called ``legs of the
manipulator'' \cite{Angeles:97}.
\subsection{Kinematic relations}
For a manipulator, the relation permitting the connection of input
values (\negr q) with output values (\negr X) is the following
\be
        F(\negr X, \negr q)=0
        \protect\label{equation:the_kinematic}
\ee
This definition can be applied to serial or parallel manipulators.
Differentiating equation (\ref{equation:the_kinematic}) with
respect to time leads to the velocity model
\be
     \negr A \negr t + \negr B \dot{\negr q} = 0
\ee
With
\beqa
     \negr t&=&\left[\begin{array}{c}
                        w \\
                        \dot{\negr c}
                     \end{array}
               \right]~For~planar~manipulators. \nonumber \\
     \negr t&=&\left[\begin{array}{c}
                       \negr w
                     \end{array}
               \right]~For~spherical~manipulators. \nonumber \\
     \negr t&=&\left[\begin{array}{c}
                      \negr w \\
                      \dot{\negr c}
                  \end{array}
               \right]~For~spatial~manipulators. \nonumber
\eeqa
Where $w$ is the scalar angular-velocity and $\dot{\negr c}$ is the
two-dimensional velocity vector of the operational point of the
moving platform for the planar manipulator. For the spherical and
the spatial manipulator, \negr w is the three-dimensional angular
velocity-vector of the moving platform. And $\dot{\negr c}$ is the
three-dimensional velocity vector of the operational point of the
moving platform for the spatial manipulator.
\par
Moreover, \negr A and \negr B are respectively the
direct-kinematics and the inverse-kinematics matrices of the
manipulator. A singularity occurs whenever \negr A or
\negr B, (or both) that can no longer be inverted. Three
types of singularities exist \cite{Gosselin:90}:
\beqa
    det(\negr A) &=& 0                       \nonumber \\
    det(\negr B) &=& 0                       \nonumber \\
    det(\negr A) &=& 0 \quad and \quad det(\negr B) = 0  \nonumber
\eeqa
\subsection{Parallel singularities}
Parallel singularities occur when the determinant of the direct
kinematics matrix \negr A vanishes. The corresponding singular
configurations are located inside the workspace. They are
particularly undesirable because the manipulator can not resist any
force and control is lost.
\subsection{Serial singularities}
Serial singularities occur when the determinant of the inverse
kinematics matrix \negr B vanishes. When the manipulator is in such
a singularity, there is a direction along which no Cartesian
velocity can be produced.
\subsection{Postures and assembling modes}
The multiple inverse kinematic solutions induce multiple postures
for each leg.
\begin{Def}
A {\em posture changing trajectory} is e\-qui\-va\-lent to a
trajectory between two inverse kinematic solutions.
\end{Def}
The multiple direct kinematic solutions induce multiple assembling
modes for the mobile platform.
\begin{Def}
An {\em assembling mode changing trajectory} is equivalent to a
trajectory between two direct kinematic solutions.
\end{Def}
\par
\begin{figure}[hbtp]
     \begin{center}
 				\epsfig{file = 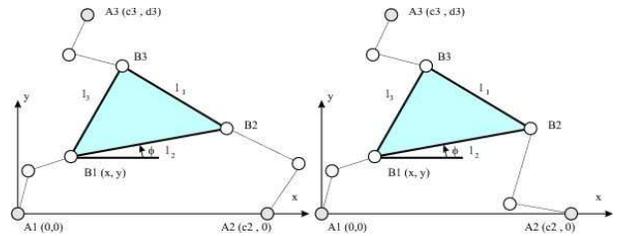, width= 80mm,height= 30mm}
        \caption{Two postures}
        \protect\label{figure:two_postures}
     \end{center}
\end{figure}
\begin{figure}[hbtp]
     \begin{center}
 				\epsfig{file = 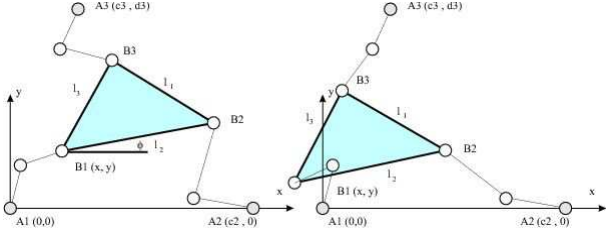, width= 80mm,height= 30mm}
        \caption{Two assembling modes}
        \protect\label{figure:two_assembling_modes}
     \end{center}
\end{figure}

As an example, the 3-RRR planar parallel manipulator (the first
joints are actuated joints), a posture changing trajectory exists
between two inverse kinematic solutions (Fig.
\ref{figure:two_postures}) and an assembling mode trajectory exits
between two direct kinematic solutions (Fig.
\ref{figure:two_assembling_modes}). In these trajectories, the
mobile platform can meet a singular configuration.
\subsection{Working Modes}
\begin{figure}[hbt]
    \begin{center}
    \begin{tabular}{cc}
       \begin{minipage}[t]{40 mm}
           \centerline{\hbox{\includegraphics[width= 40mm,height= 45mm]{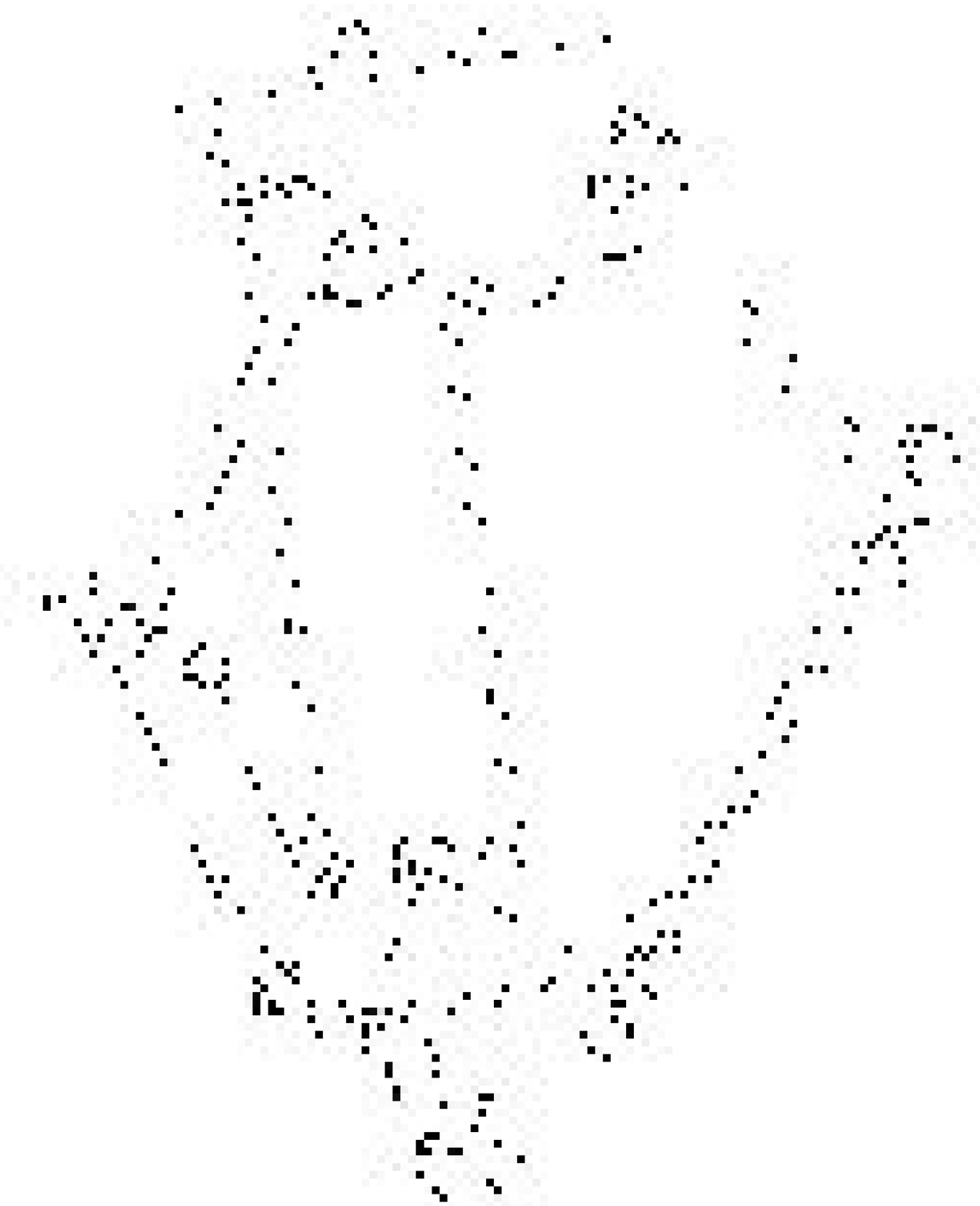}}}
           \caption{The Delta manipulator}
           \protect\label{figure:Delta_manipulateur}
       \end{minipage} &
       \begin{minipage}[t]{40 mm}
           \centerline{\hbox{\includegraphics[width= 40mm,height= 40mm]{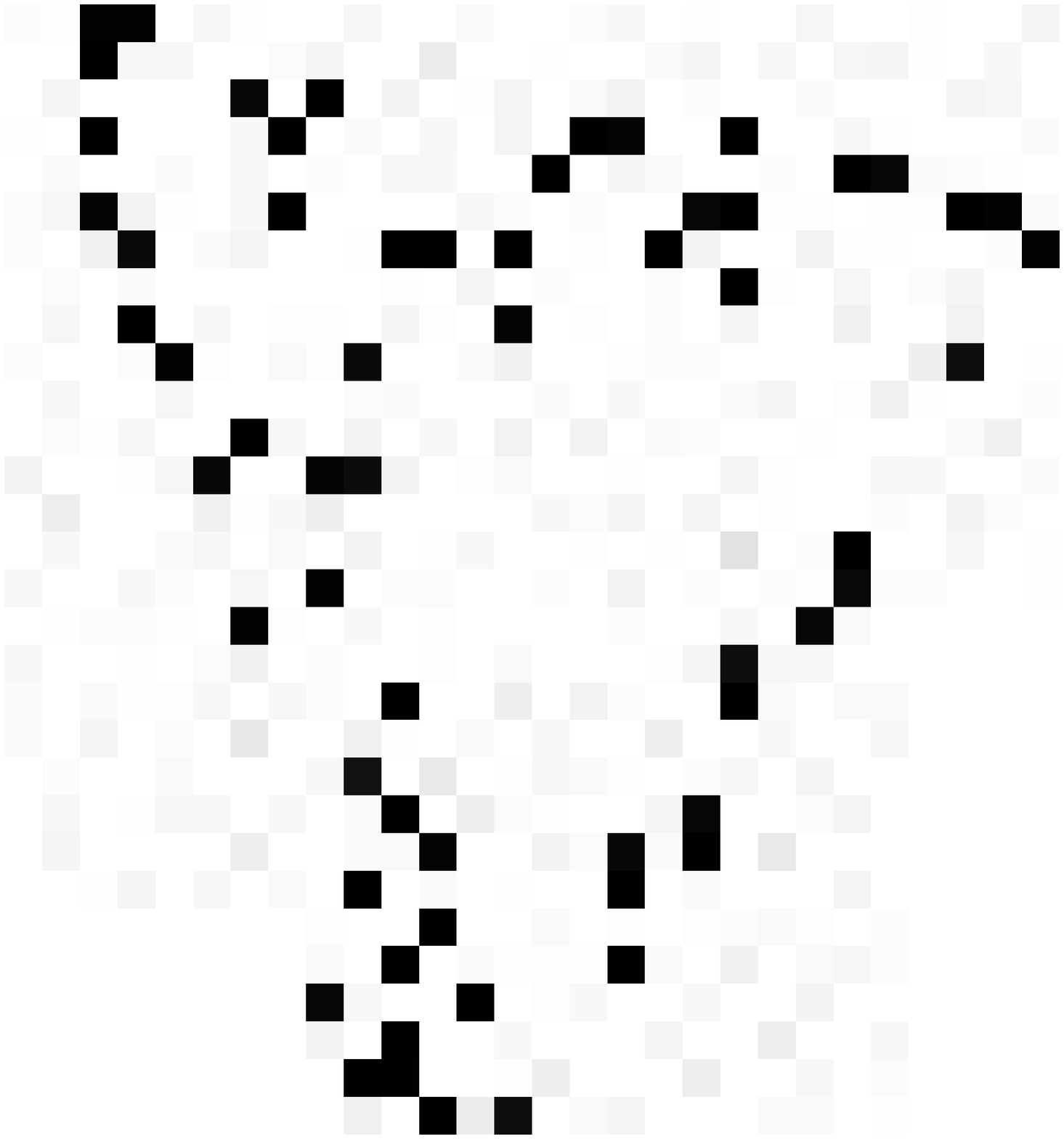}}}
           \caption{The Hexa manipulator}
           \protect\label{figure:Hexa_manipulateur}
       \end{minipage}
    \end{tabular}
    \end{center}
\end{figure}

The {\em working modes} are defined for fully parallel manipulators
(Def. \ref{Definition:Fully_Parallel_Manipulator}). From this
definition, the inverse-kinematic matrix is always diagonal. For a
manipulator with $n$ degrees of freedom, the inverse kinematic
matrix \negr B is like in eq.~(\ref{equation:the_B_matrix}). Each
term $\negr B_{jj}$ is associated with one leg. Its vanishing
induces the apparition of a serial singularity.
\be
\negr B = \left[\begin{array}{ccccc}
               \negr B_{11} &    0   & \cdots       & \cdots &    0       \\
               0            & \ddots &              & \ddots & \vdots     \\
               \vdots       &        & \negr B_{jj} &        & \vdots     \\
               \vdots       & \ddots &              & \ddots &    0       \\
               0            & \cdots & \cdots       &   0    & \negr B_{nn}
                \end{array}
          \right]
          \label{equation:the_B_matrix}
\ee
Let $W$ be the reachable workspace, that is, the set of all
positions and orientations reachable by the moving platform
(\cite{Kumar:92} and \cite{Pennock:93}). Let $Q$ be the reachable
joint space, that is, the set of all joint vectors reachable by
actuated joints.
\begin{Def}
A {\em working mode}, noted $Mf_i$, is the set of postures for
which the sign of $\negr B_{jj}$ ($j = 1$ to $n$) does not change
and $\negr B_{jj}$ does not vanish.
\be
Mf_i = \left\{ (\negr X, \negr q) \in W \cdot Q \setminus
              \begin{array}{c}
                  sign(\negr B_{jj})=constant \\
                  for (j=1~to~n) \\
                  and~det(\negr B) \neq 0
              \end{array}
       \right\}
\ee
Therefore, the set of working modes ($Mf=\left\{Mf_i\right\}$,
$i\in~I$) is obtained while using all permutations of sign of each
term $\negr B_{jj}$.
\end{Def}
The Cartesian product of $W$ by $Q$ is noted $W \cdot Q$. According
to the joint limit values, all working modes do not necessarily
exist. Changing working mode is equivalent to changing the posture
of one or several given legs. The working modes are defined in $W
\cdot Q$ because the terms $\negr B_{jj}$ depend on both \negr X and \negr q.
\begin{Theo}
The working modes separate inverse kinematic solutions if and only
if the legs are not cuspidal (see \cite{Wenger:93}).
\end{Theo}
\begin{Pro}
If one leg is cuspidal then this leg can make a changing posture
trajectory without meeting a serial singularity. In this case no
$\negr B_{jj}$ vanishes during this trajectory. Reciprocally, if no
leg is cuspidal, then the changing posture trajectory of one leg
induces that some $B_{jj}$ can vanish.
\end{Pro}
In this study, the legs are not cuspidal so that the working modes
allow the separation of the inverse kinematic solutions. The list
of the most current noncuspidal serial chains is given in
\cite{Wenger:93}.
\begin{Exa}
For the robot Delta \cite{Clavel:88}, a 3-dof manipulator
(Fig.~\ref{figure:Delta_manipulateur}), there are 8 working modes
($3$ legs and $2$ postures for each leg, with $2^3=8$ working
modes). And for the Hexa robot \cite{Pierrot:91}, a 6-dof
manipulator (Fig.~\ref{figure:Hexa_manipulateur}), there are $64$
working modes ($2^6=64$). For these manipulators, the serial
singularities occur when one or more legs are outstretched.
\end{Exa}
\subsection{Notion of aspect for fully parallel manipulators: General definition}
The notion of aspect was introduced by \cite{Borrel:86} to cope
with the existence of multiple inverse kinematic solutions in
serial manipulators. Recently, the notion of aspect was defined for
parallel manipulators with only one inverse kinematic solution
\cite{Wenger:97} to cope with the existence of multiple direct
kinematic solutions.
\par
In this section, the  notion of aspect is redefined formally for
fully parallel manipulators with multiple inverse and direct
kinematic solutions.
\begin{Def}
\label{definition:Aspect}
The generalized aspects $\negr A_{ij}$ are defined as the maximal
sets in $W \cdot Q$ so that
\end{Def}
\begin{itemize}
\item $\negr A_{ij} \subset W \cdot Q$;
\item $\negr A_{ij}$ is connected;
\item $\negr A_{ij} = \left\{
                   (\negr X, \negr q) \in Mf_i \setminus det(\negr A) \neq 0
                  \right\}$
\end{itemize}
In other words, the generalized aspects $\negr A_{ij}$ are the
maximal singularity-free domains of the Cartesian product of the
reachable workspace with the reachable joint space.
\begin{Def}
The projection of the generalized aspects in the workspace yields
the parallel aspects ${\bf WA}_{ij}$ so that
\end{Def}
\begin{itemize}
\item ${\bf WA}_{ij} \subset W$;
\item ${\bf WA}_{ij}$ is connected.
\end{itemize}
The parallel aspects are the maximal singularity-free domains in
the workspace for one given working mode.
\begin{Def}
The projection of the  generalized aspects in the joint space
yields the serial aspects ${\bf QA}_{ij}$ so that
\end{Def}
\begin{itemize}
\item ${\bf QA}_{ij} \subset Q$;
\item ${\bf QA}_{ij}$ is connected.
\end{itemize}
The serial aspects are the maximal singularity-free domains in the
joint space for one given working mode.
\section{A Two-DOF Closed-Chain Manipulator}
\begin{figure}[hbt]
    \begin{center}
 				\epsfig{file = 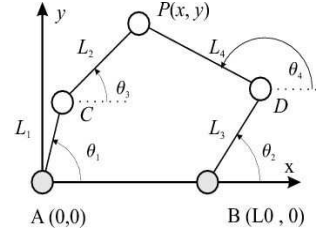, width= 40mm,height= 30mm}
        \caption{A two-dof closed-chain manipulator}
        \protect\label{figure:manipulateur_general}
    \end{center}
\end{figure}

For more legibility, a planar manipulator is used as illustrative
example in this paper. This is a five-bar, revolute
($R$)-closed-loop linkage, as displayed in
Fig.~\ref{figure:manipulateur_general}. The actuated joint
variables are $\theta_1$ and $\theta_2$, while the Output values
are the ($x$, $y$) coordinates of the revolute center $P$. The
passive joints will always be assumed unlimited in this study.
Lengths $L_0$, $L_1$, $L_2$, $L_3$, and $L_4$ define the geometry
of this manipulator entirely. We assume here the dimensions
$L_0~=~9$, $L_1~=~8$, $L_2~=~5$, $L_3~=~5$ and $L_4~=8~$, in
certain units of length that we need not specify.
\subsection{Kinematic Relations}
The velocity $\dot{\bf p}$ of point $P$, of position vector \negr p,
can be obtained in two different forms, depending on the direction in
which the loop is traversed, namely,
\begin{subequations}
\be 
        \dot{\bf p}= \dot{\bf c} + \dot{\theta}_3 {\bf E}
                     ({\bf p}  - {\bf c} )
        \protect\label{equation:kinematic_1}
\ee
\be
        \dot{\bf p}= \dot{\bf d} + \dot{\theta}_4 {\bf E}
                     ({\bf p}  - {\bf d}
        \protect\label{equation:kinematic_2}
\ee
with matrix ${\bf E}$ defined as
\begin{eqnarray}
{\bf E}= \left[\begin{array}{cc}
              0 & -1 \\
              1 &  0
             \end{array}
        \right]                  \nonumber
\end{eqnarray}
\end{subequations}
and \negr c and \negr d denoting the position vectors, in the frame
indicated in Fig.~\ref{figure:manipulateur_general}, of points $C$
and $D$, respectively. Furthermore, note that $\dot{\negr c}$ and
$\dot{\negr d}$ are given by
\begin{eqnarray}
\dot{\negr c}= \dot{\theta}_1 {\bf E}\negr c, \quad\dot{\negr d}=
\dot{\theta}_2 {\bf E}(\negr d - \negr b)   \nonumber
\end{eqnarray}
We would like to eliminate the two idle joint rates $\dot{\theta}_3$ and
$\dot{\theta}_4$ from eqs.(\ref{equation:kinematic_1}) and
(\ref{equation:kinematic_2}), which we do upon dot-multiplying the
former by $\negr p-\negr c$ and the latter by $\negr p-\negr d$, thus
obtaining
\begin{subequations}
\be
      ({\bf p - c})^T \dot{\bf p} = ({\bf p} - {\bf c})^T \dot{\bf c}
      \protect\label{equation:kinematic_3}
\ee
\be
      ({\bf p - d})^T \dot{\bf p} = ({\bf p} - {\bf d})^T \dot{\bf d}
      \protect\label{equation:kinematic_4}
\ee
\end{subequations}
Equations (\ref{equation:kinematic_3}) and
(\ref{equation:kinematic_4}) can now be cast in vector form, namely,
\begin{subequations}
\be
{\bf A} \dot{\negr p}={\bf B \dot{\gbf\theta}}\label{e:Adp=Bdth}
\ee
with $\dot{\gbf{\theta}}$ defined as the vector of actuated joint
rates, of components $\dot{\theta}_1$ and $\dot{\theta}_2$.
Moreover ${\bf A}$  and \negr B are, respectively, the
direct-kinematics and the inverse-kinematics matrices of the
manipulator,  defined as
\be
{\bf A}= \left[\begin{array}{c}
                ({\bf p}  - {\bf c})^T \\
                ({\bf p}  - {\bf d})^T
              \end{array}
         \right]
       \protect\label{equation:jacobian_matrices_A}
\ee
and
\be
{\bf B}=  \left[\begin{array}{cc}
                   L_1 L_2 \sin(\theta_3 - \theta_1) \!\!\!&
                   0                                 \\
                   0                                 &
                   \!\!\!L_3 L_4 \sin(\theta_4 - \theta_2)
                \end{array}
           \right]
       \protect\label{equation:jacobian_matrices_B}
\ee
\end{subequations}
\subsection{Parallel singularities}
\begin{figure}[hbt]
    \begin{center}
 				\epsfig{file = 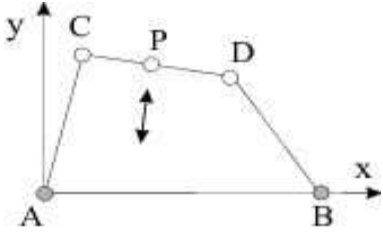, width= 50mm,height= 30mm}
        \caption{Example of parallel singularity}
        \protect\label{figure:parallel_singularity}
    \end{center}
\end{figure}

For the manipulator studied, the parallel singularities occur
whenever the points $C$, $D$, and $P$ are aligned
(\ref{figure:parallel_singularity}). Manipulator postures whereby
$\theta_3-\theta_4= k\pi$ denote a singular matrix \negr A, and
hence, define the boundary of the Joint space of the manipulator.
\subsection{Serial singularities}
\begin{figure}[hbtp]
    \begin{center}
 				\epsfig{file = 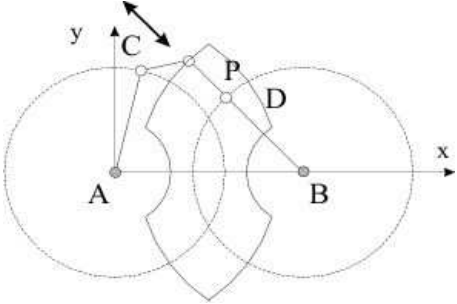, width= 60mm,height= 40mm}
        \caption{Example of serial singularity}
        \protect\label{figure:serial_singularity}
    \end{center}
\end{figure}

For the manipulator at hand, the serial singularity occur whenever
the points $A$, $C$, and $P$ or the points $B$, $D$, and $P$ are
aligned (\ref{figure:serial_singularity}). Manipulator postures
whereby $\theta_3-\theta_1= k\pi$ or $\theta_4-\theta_2= k\pi$
denote a singular matrix \negr B, and hence, define the boundary of
the Cartesian workspace of the manipulator.
\subsection{The Working Mode}
\begin{figure}[!hbt]
    \begin{center}
 				\epsfig{file = 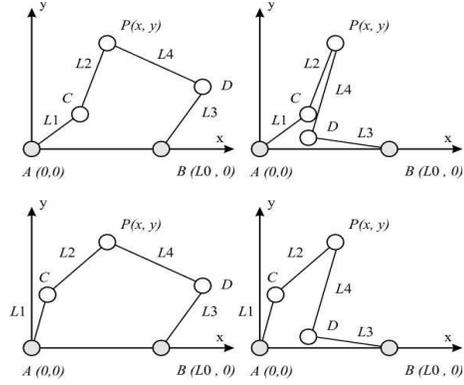, width= 60mm,height= 50mm}
        \caption{The four working modes}
        \protect\label{figure:working_mode}
    \end{center}
\end{figure}

\begin{figure}[!hbt]
    \begin{center}
    \begin{tabular}{cc}
       \begin{minipage}[t]{40 mm}
           \centerline{\hbox{\includegraphics[width= 20mm,height= 20mm]{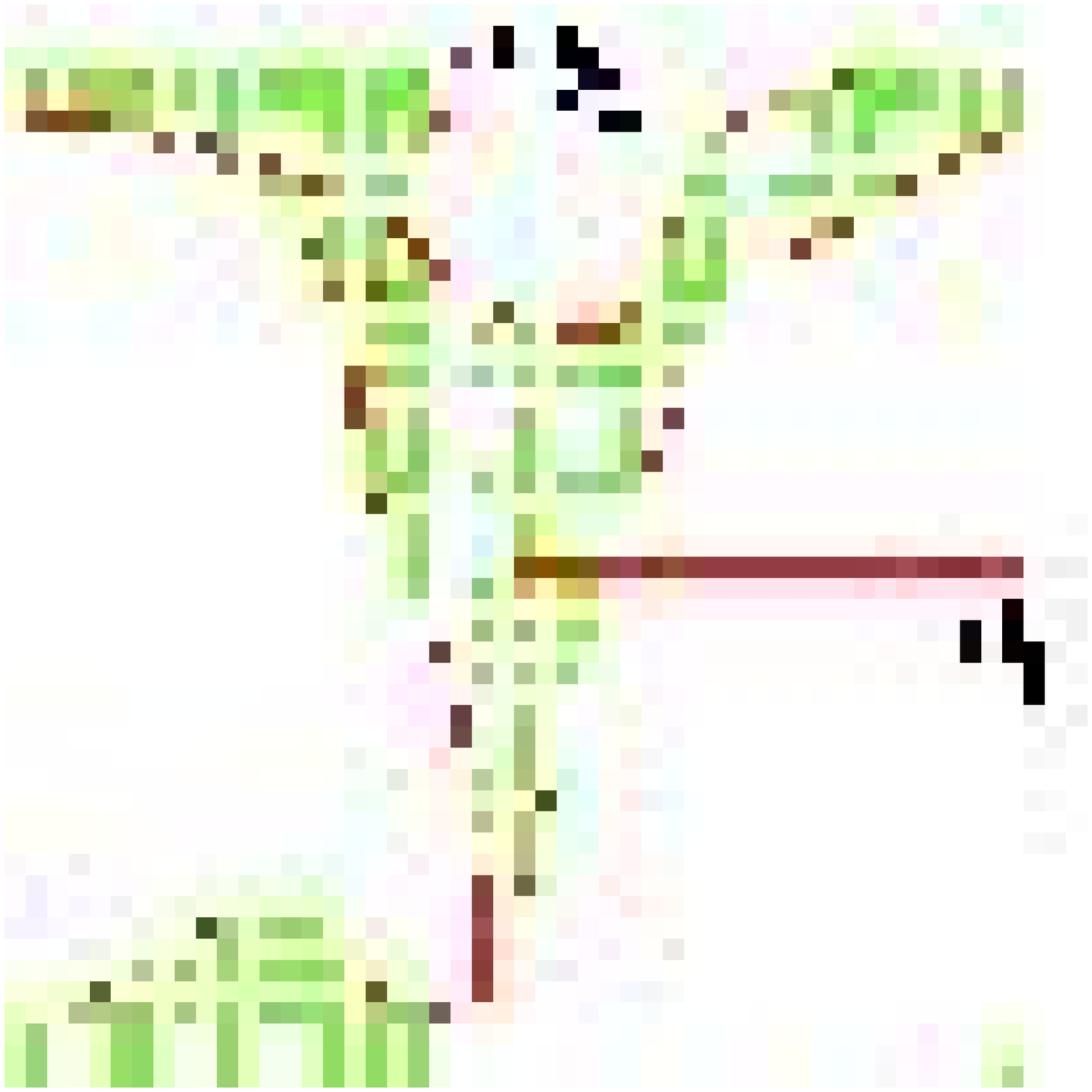}
                             \includegraphics[width= 20mm,height= 20mm]{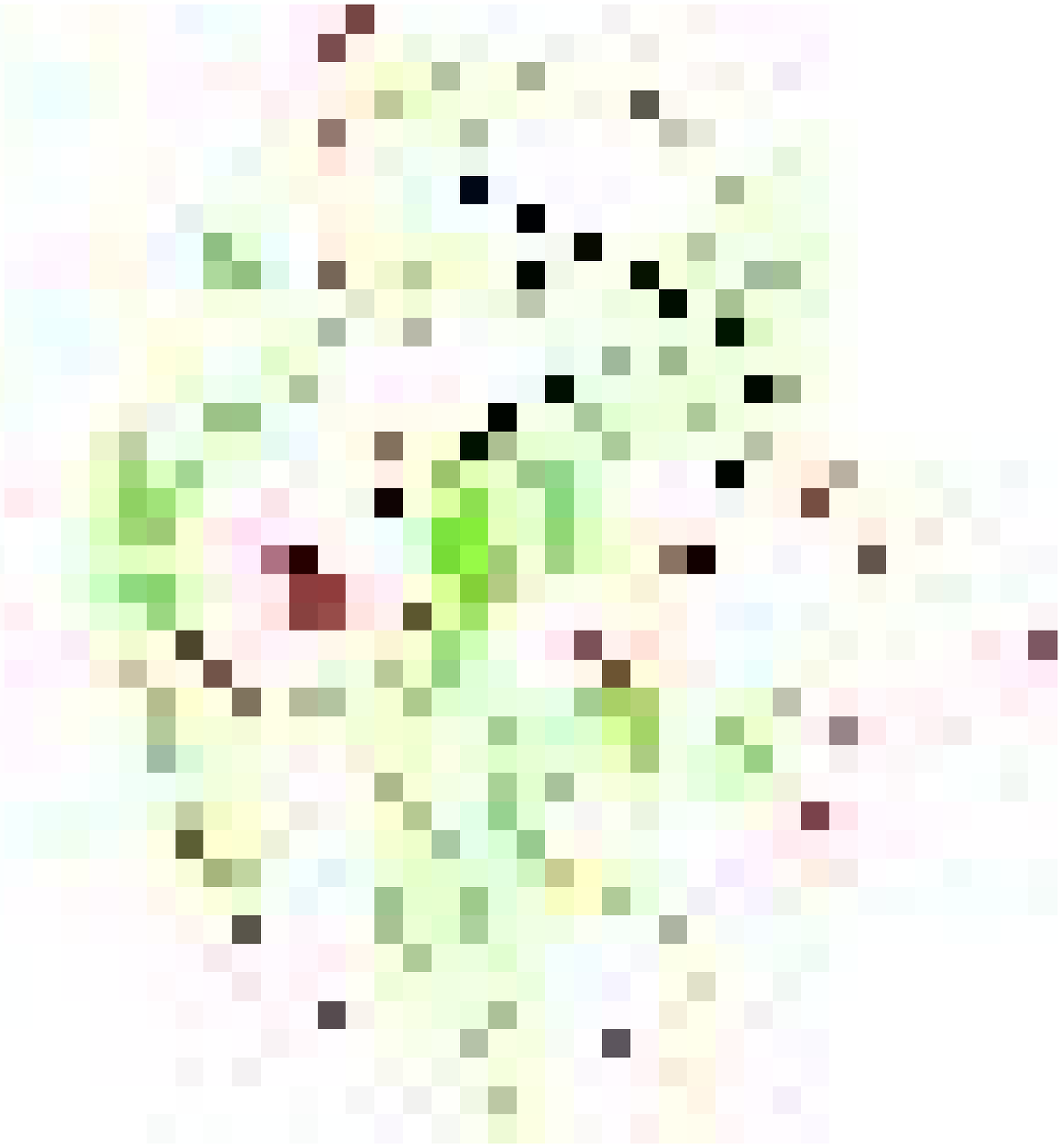}}}
           \caption{$Mf_1$}
           \label{figure:Mf_1}
       \end{minipage} &
       \begin{minipage}[t]{40 mm}
           \centerline{\hbox{\includegraphics[width= 20mm,height= 20mm]{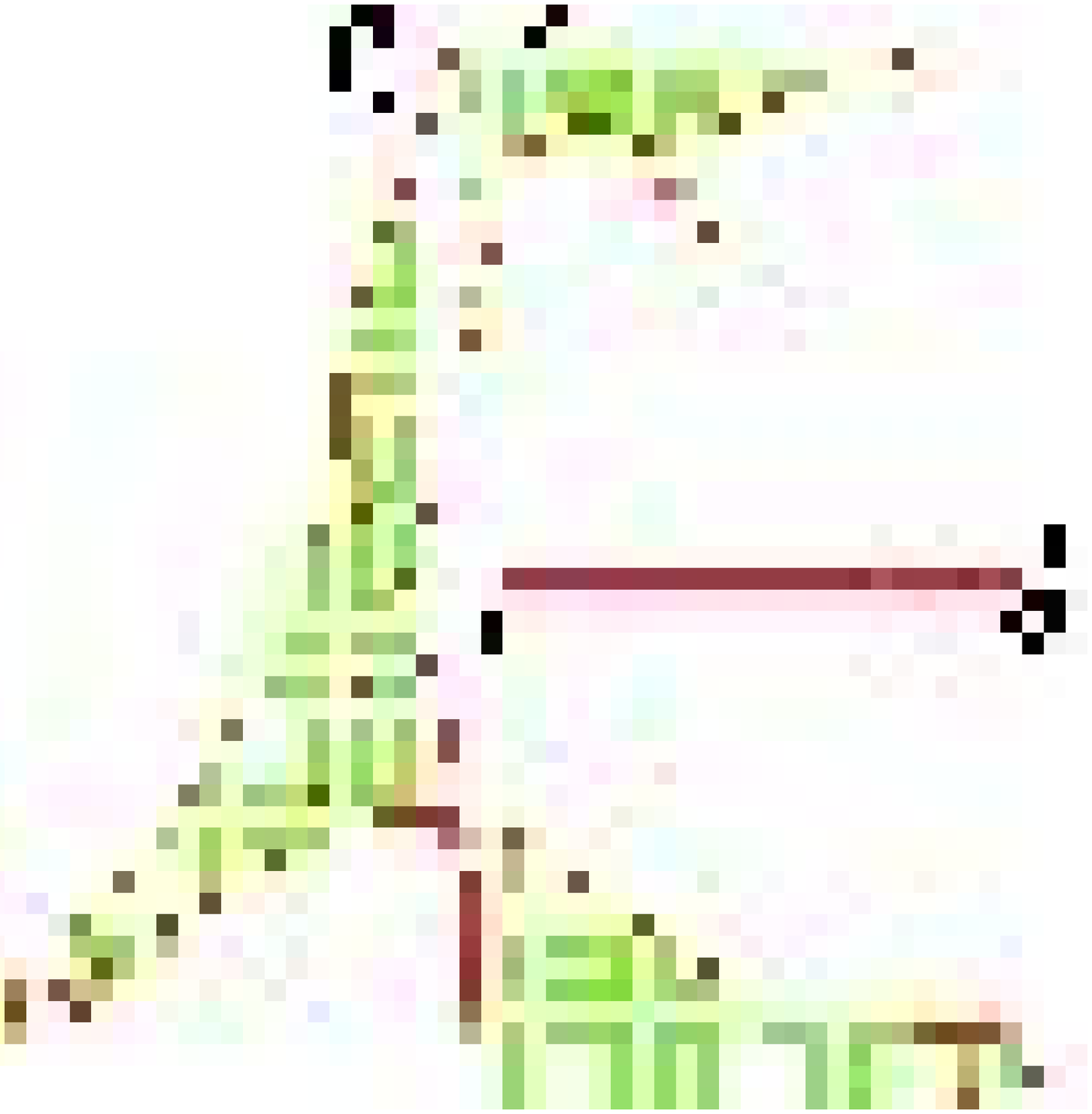}
                             \includegraphics[width= 20mm,height= 20mm]{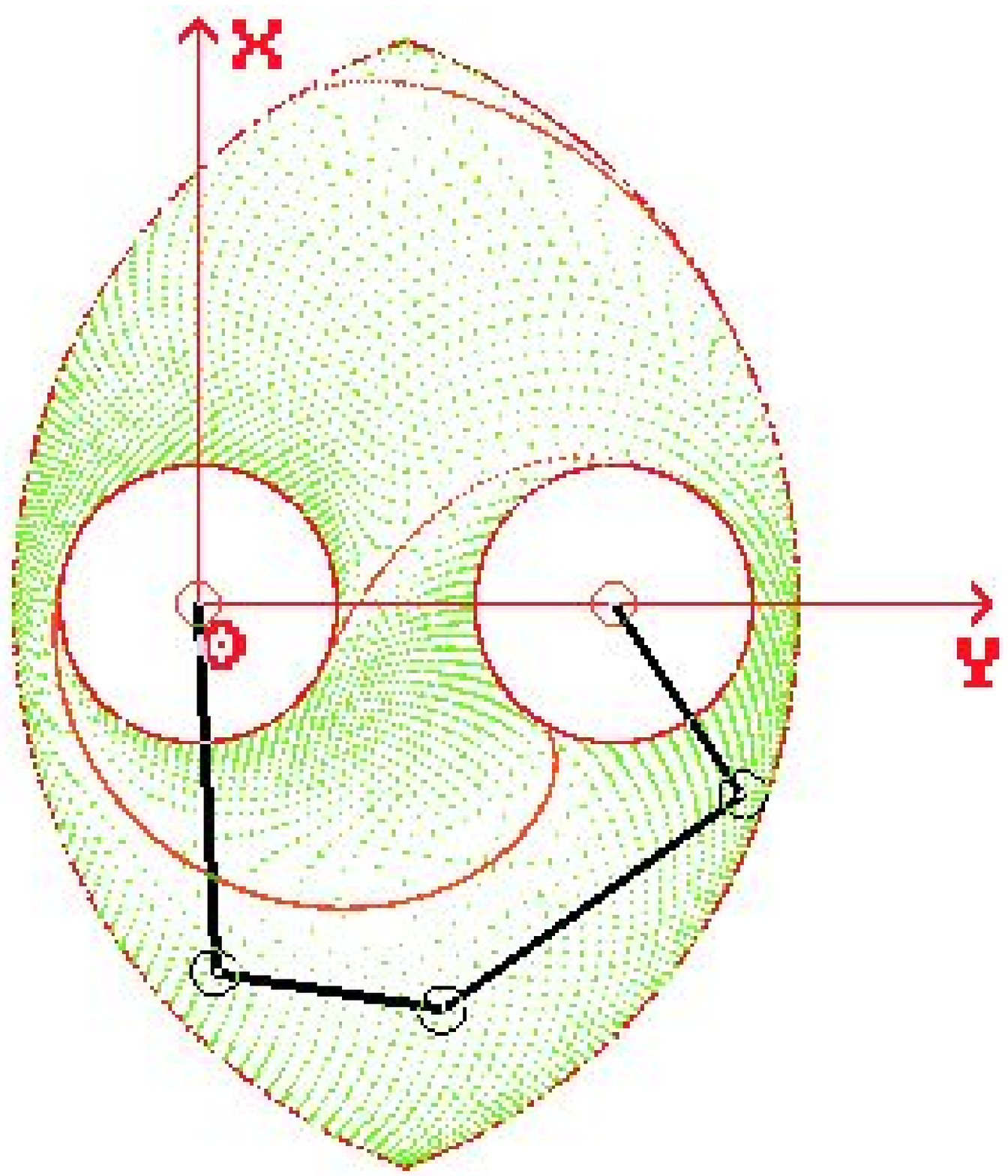}}}
           \caption{$Mf_2$}
           \label{figure:Mf_2}
       \end{minipage} \\ \\
       \begin{minipage}[t]{40 mm}
           \centerline{\hbox{\includegraphics[width= 20mm,height= 20mm]{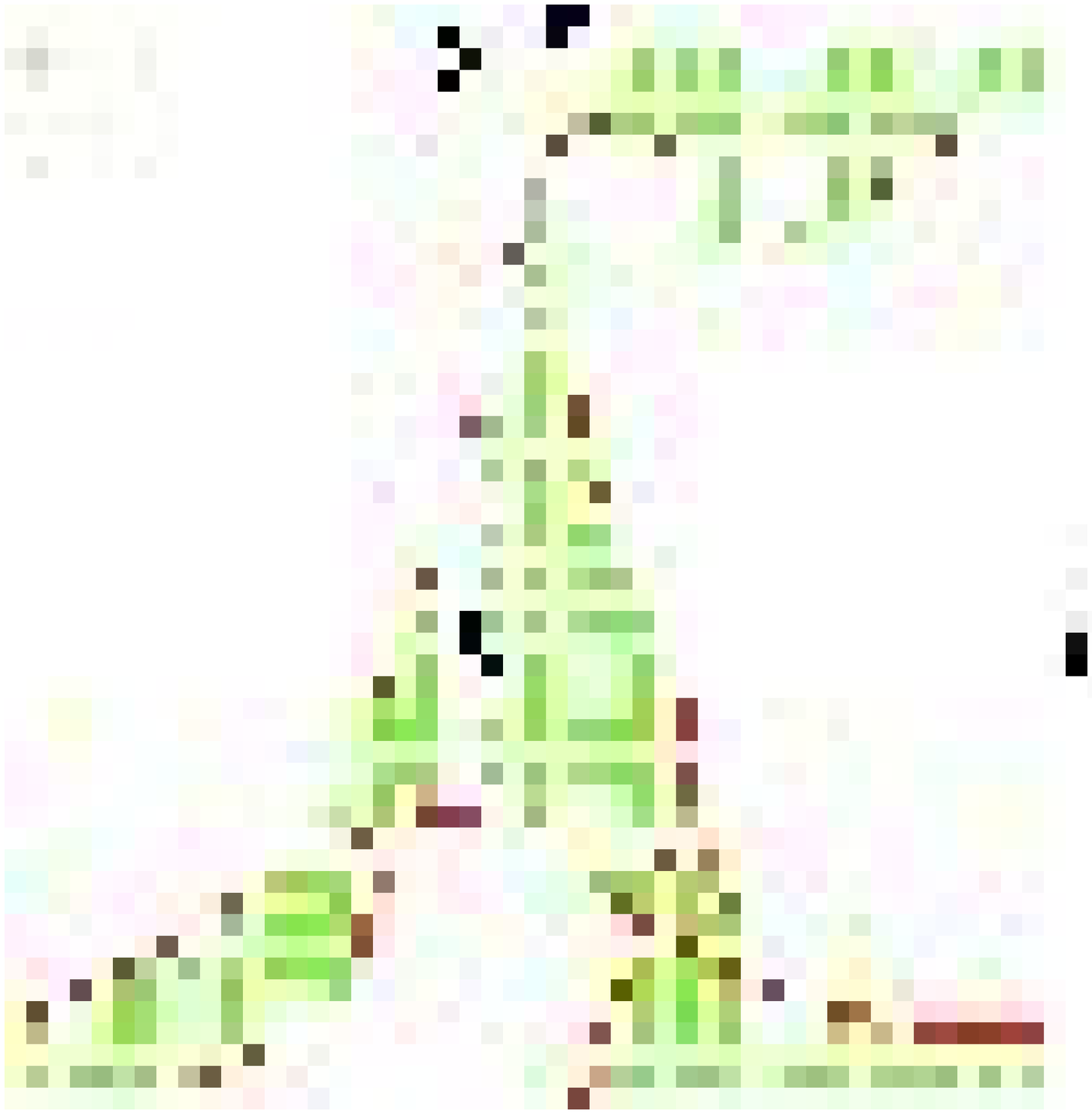}
                             \includegraphics[width= 20mm,height= 20mm]{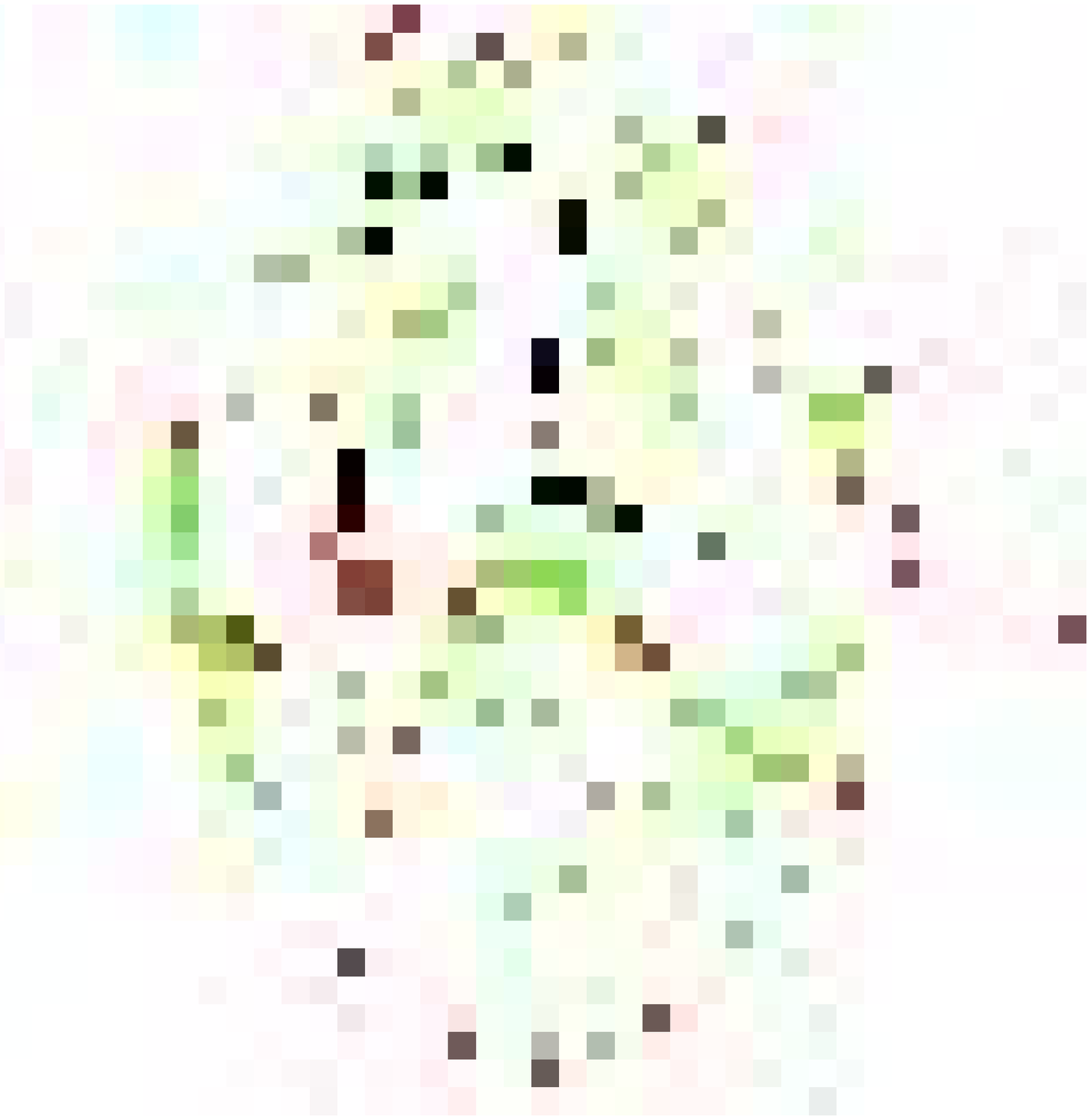}}}
           \caption{$Mf_3$}
           \label{figure:Mf_3}
       \end{minipage} &
       \begin{minipage}[t]{40 mm}
           \centerline{\hbox{\includegraphics[width= 20mm,height= 20mm]{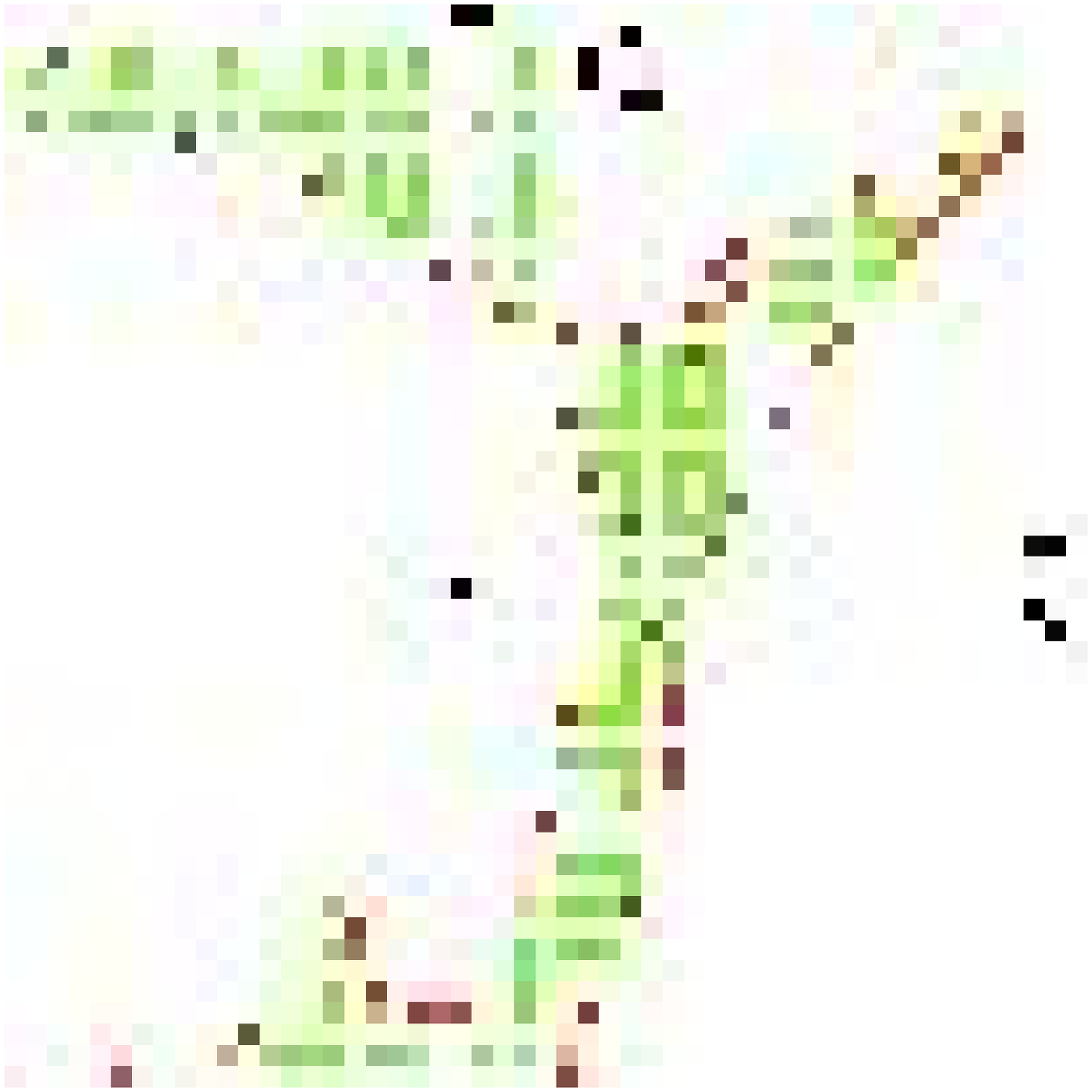}
                             \includegraphics[width= 20mm,height= 20mm]{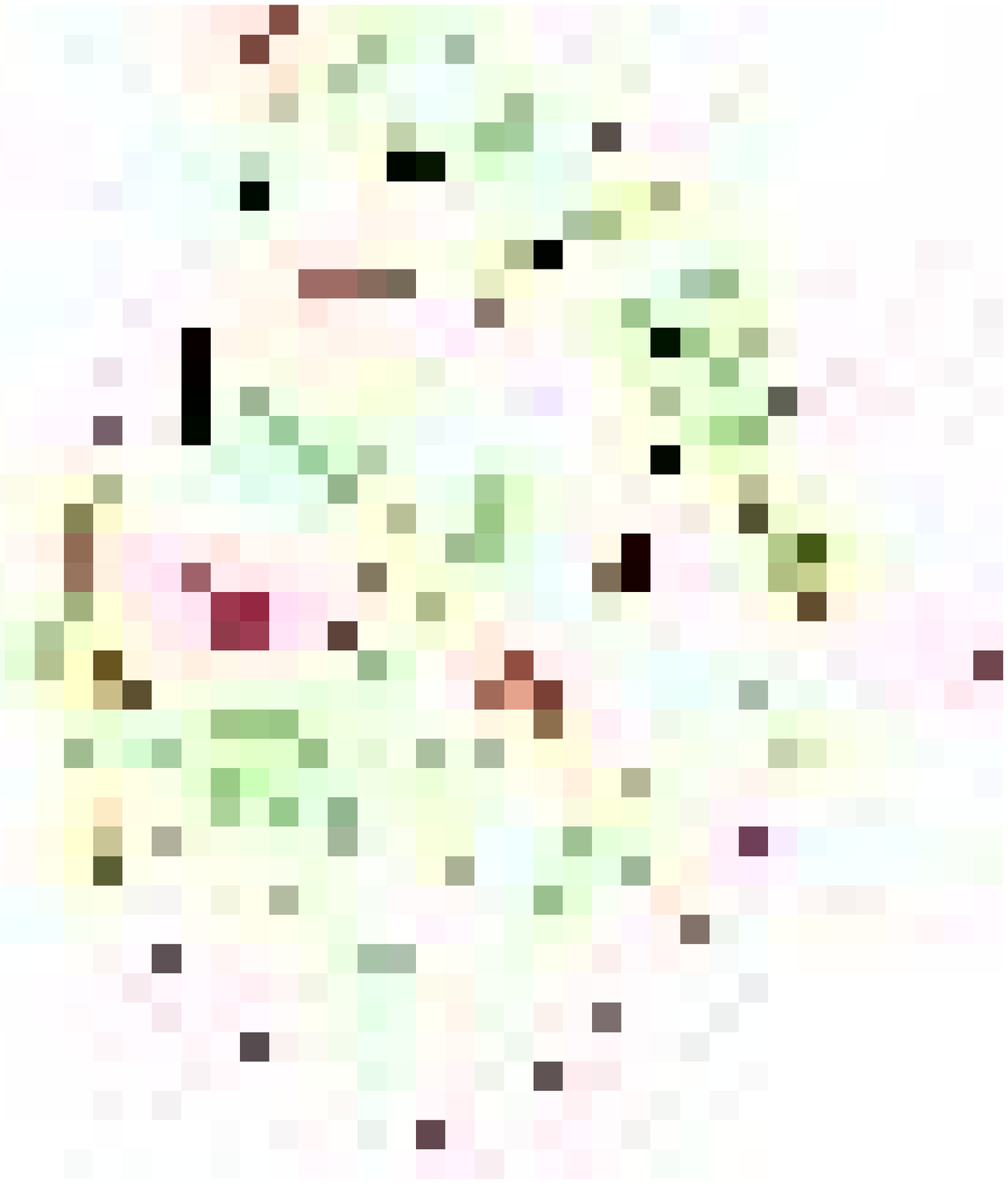}}}
           \caption{$Mf_4$}
           \label{figure:Mf_4}
       \end{minipage}
    \end{tabular}
    \end{center}
\end{figure}

The manipulator under study has a diagonal inverse-kinematics
matrix \negr B, as shown in
eq.~(\ref{equation:jacobian_matrices_B}). There are four working
modes, as depicted in Fig.~\ref{figure:working_mode}. The different
working modes in the Cartesian workspace and in the Joint space are
displayed in figures \ref{figure:Mf_1}, \ref{figure:Mf_2},
\ref{figure:Mf_3} and \ref{figure:Mf_4}.
\subsection{The generalized aspects}
\begin{figure}[hbt]
    \begin{center}
    \begin{tabular}{cc}
       \begin{minipage}[t]{40 mm}
           \centerline{\hbox{\includegraphics[width= 20mm,height= 20mm]{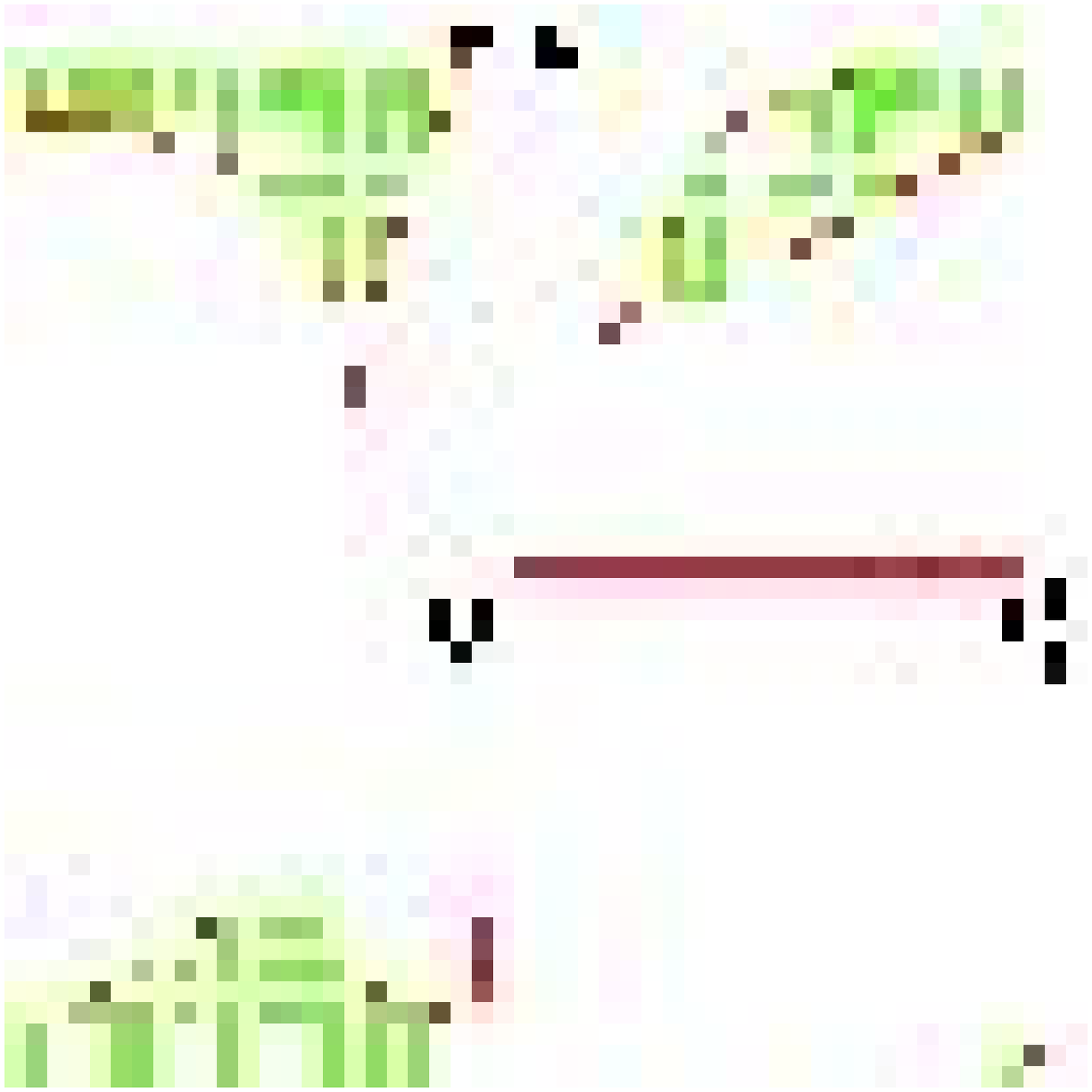}
                             \includegraphics[width= 20mm,height= 20mm]{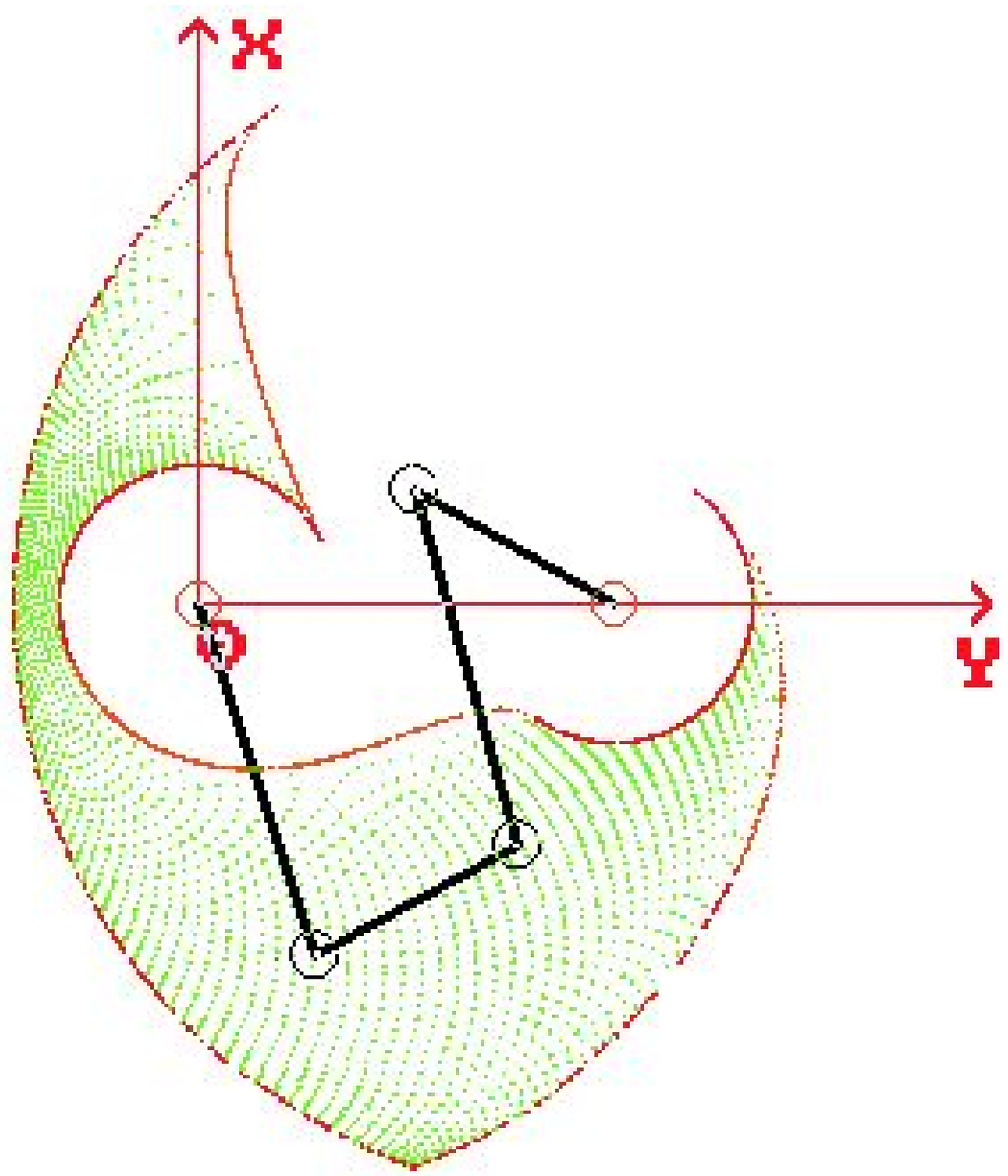}}}
           \caption{Aspect 1}
           \label{figure:Aspect_1}
       \end{minipage} &
       \begin{minipage}[t]{42 mm}
           \centerline{\hbox{\includegraphics[width= 20mm,height= 20mm]{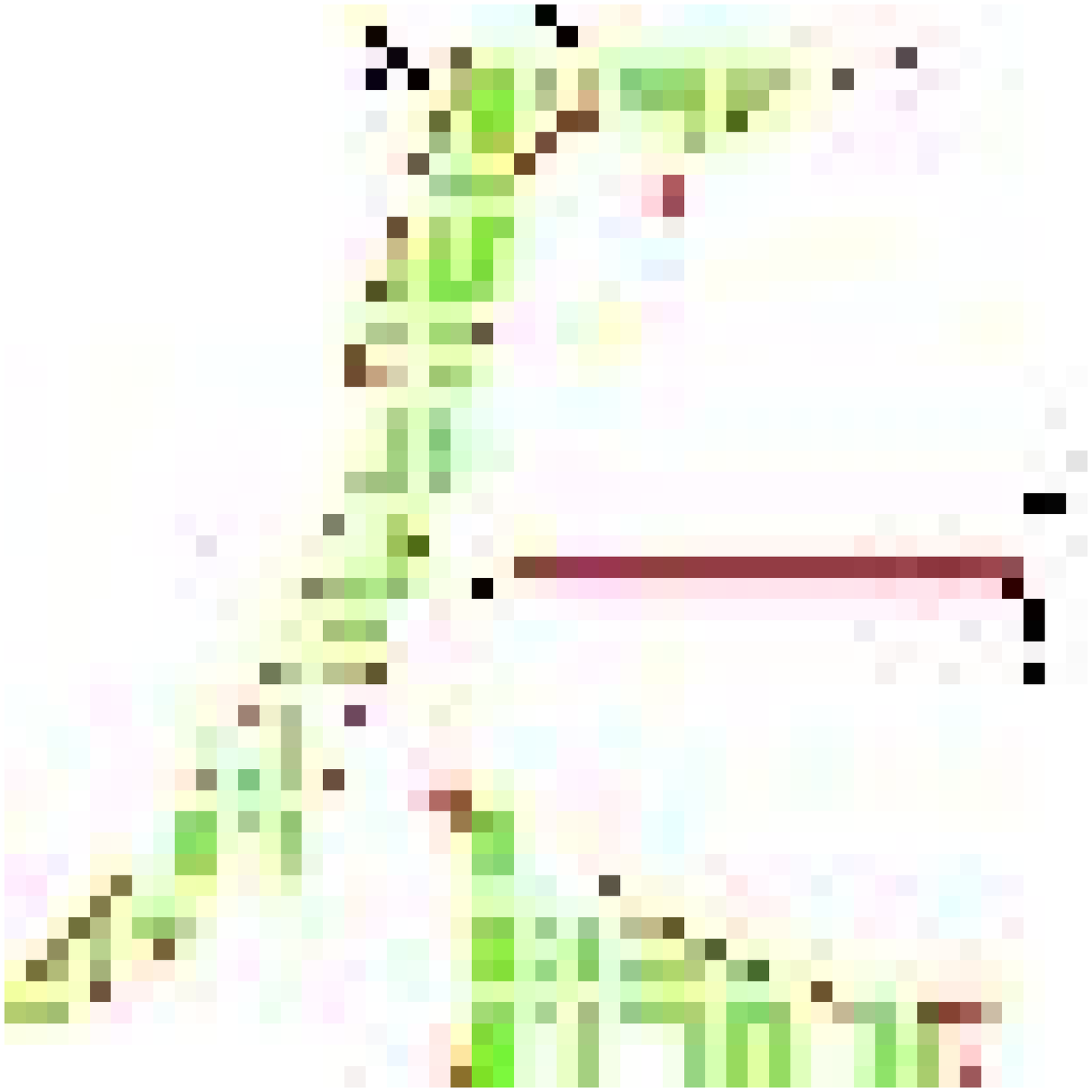}
                             \includegraphics[width= 20mm,height= 20mm]{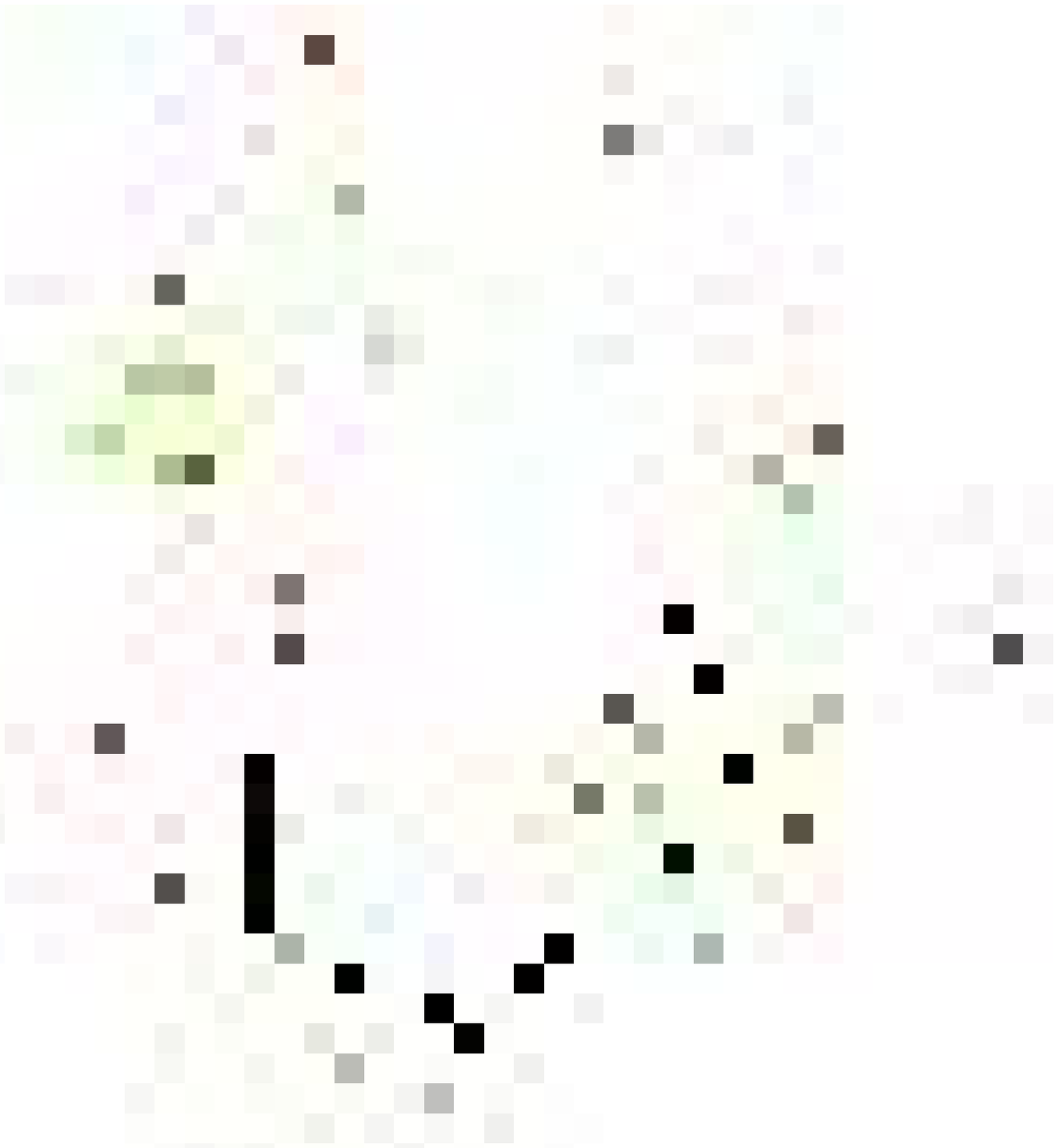}}}
           \caption{Aspects 2}
           \label{figure:Aspect_2}
       \end{minipage} \\ \\
       \begin{minipage}[t]{40 mm}
           \centerline{\hbox{\includegraphics[width= 20mm,height= 20mm]{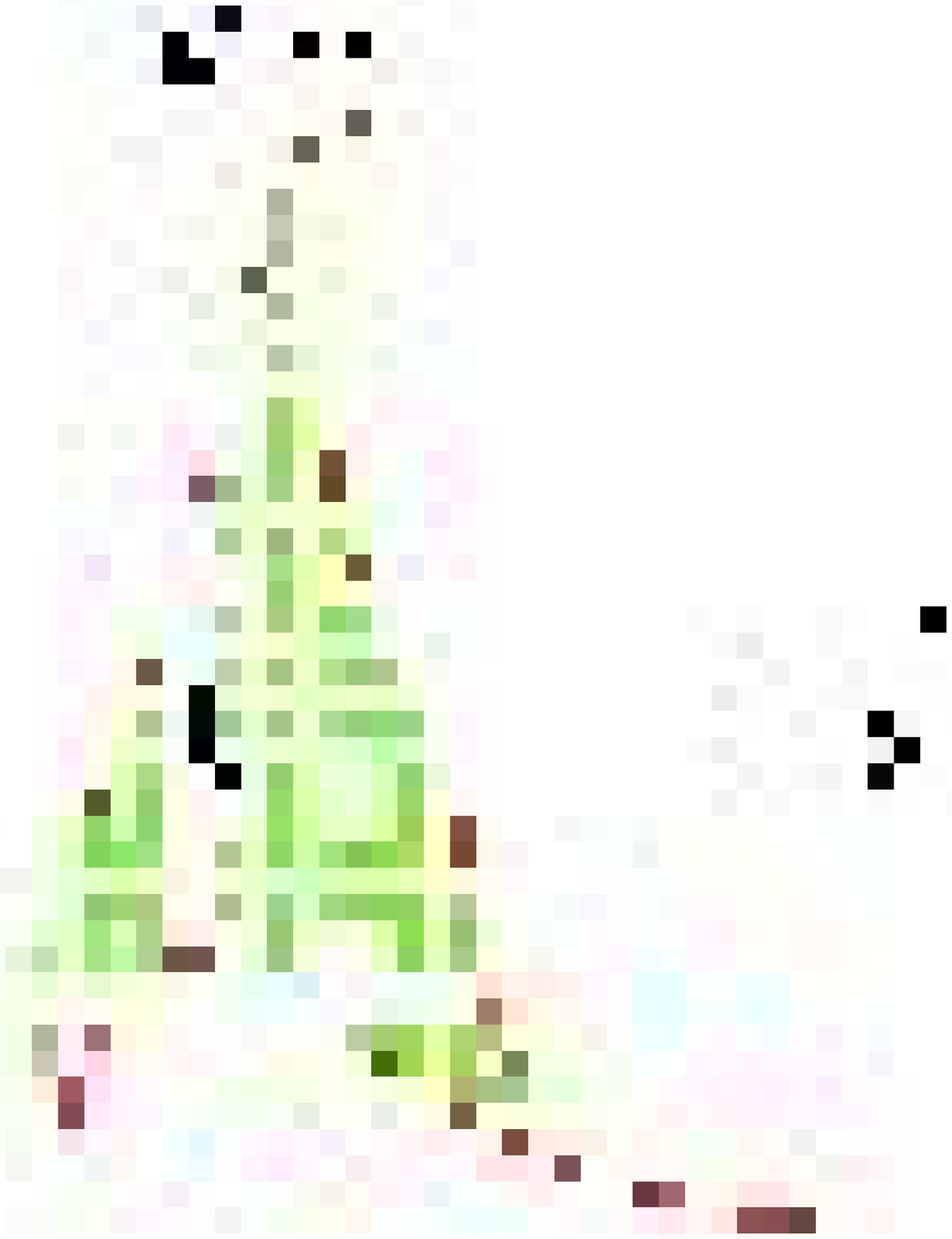}
                             \includegraphics[width= 20mm,height= 20mm]{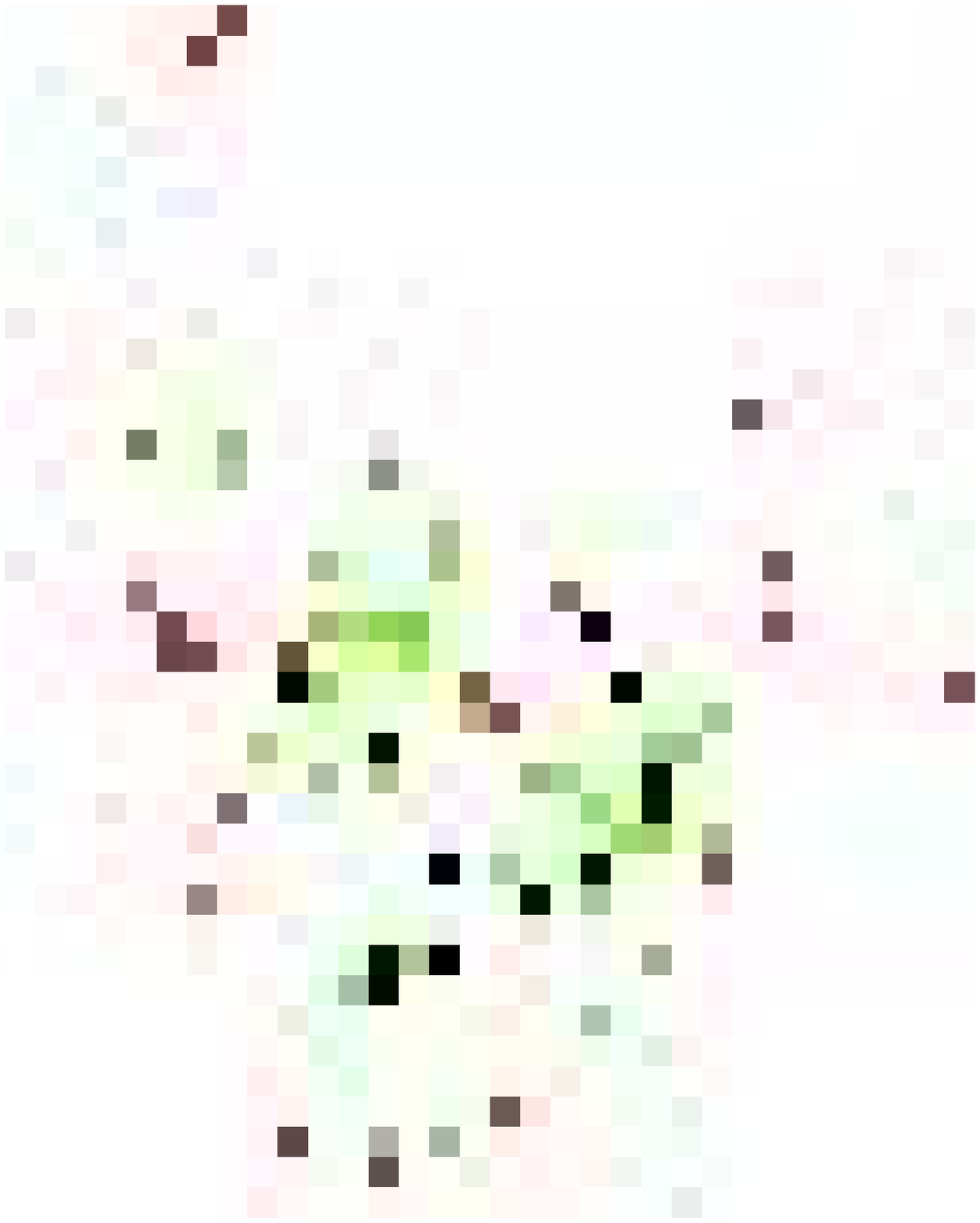}}}
           \caption{Aspect 3}
           \label{figure:Aspect_3}
       \end{minipage} &
       \begin{minipage}[t]{42 mm}
           \centerline{\hbox{\includegraphics[width= 20mm,height= 20mm]{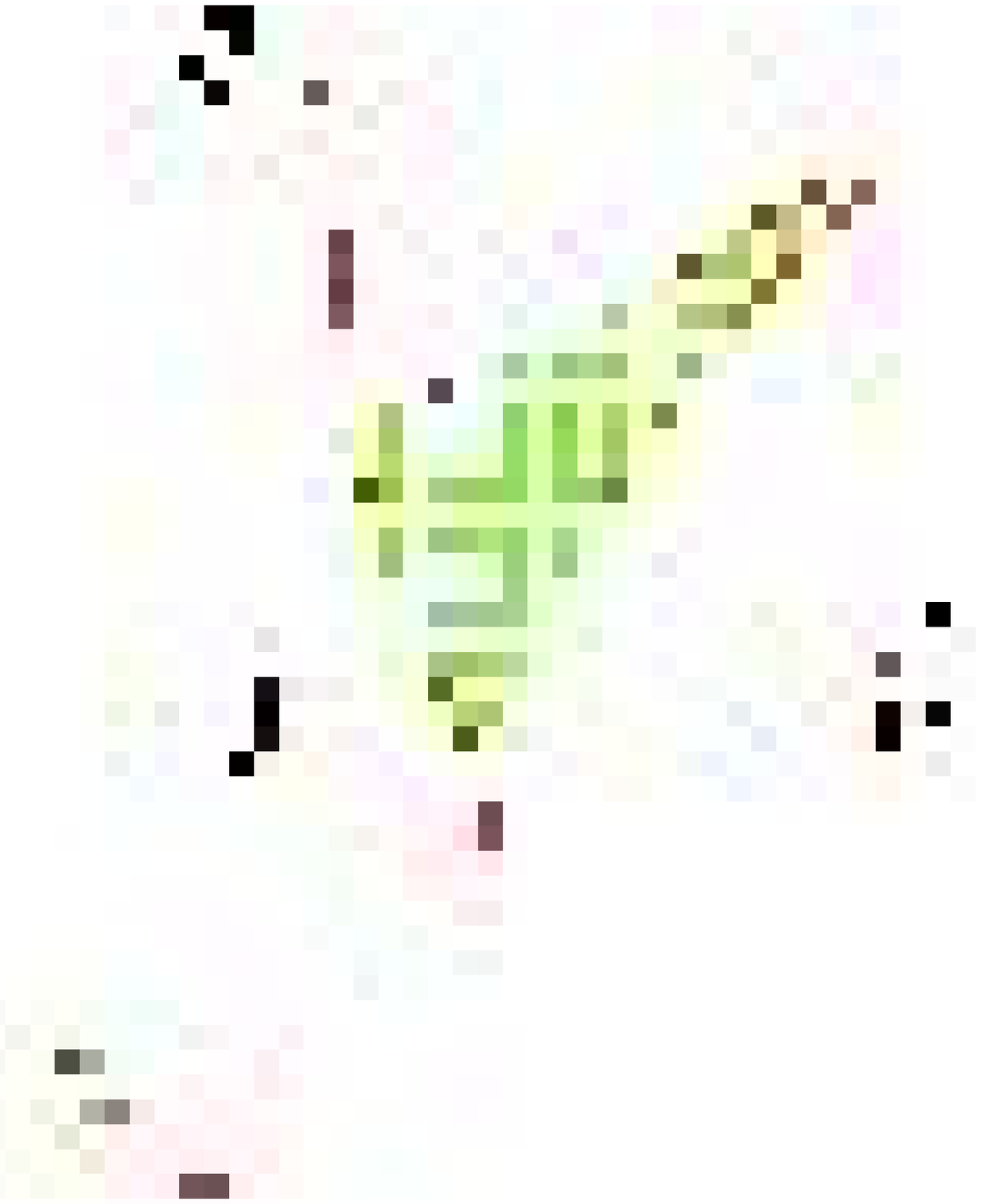}
                             \includegraphics[width= 20mm,height= 20mm]{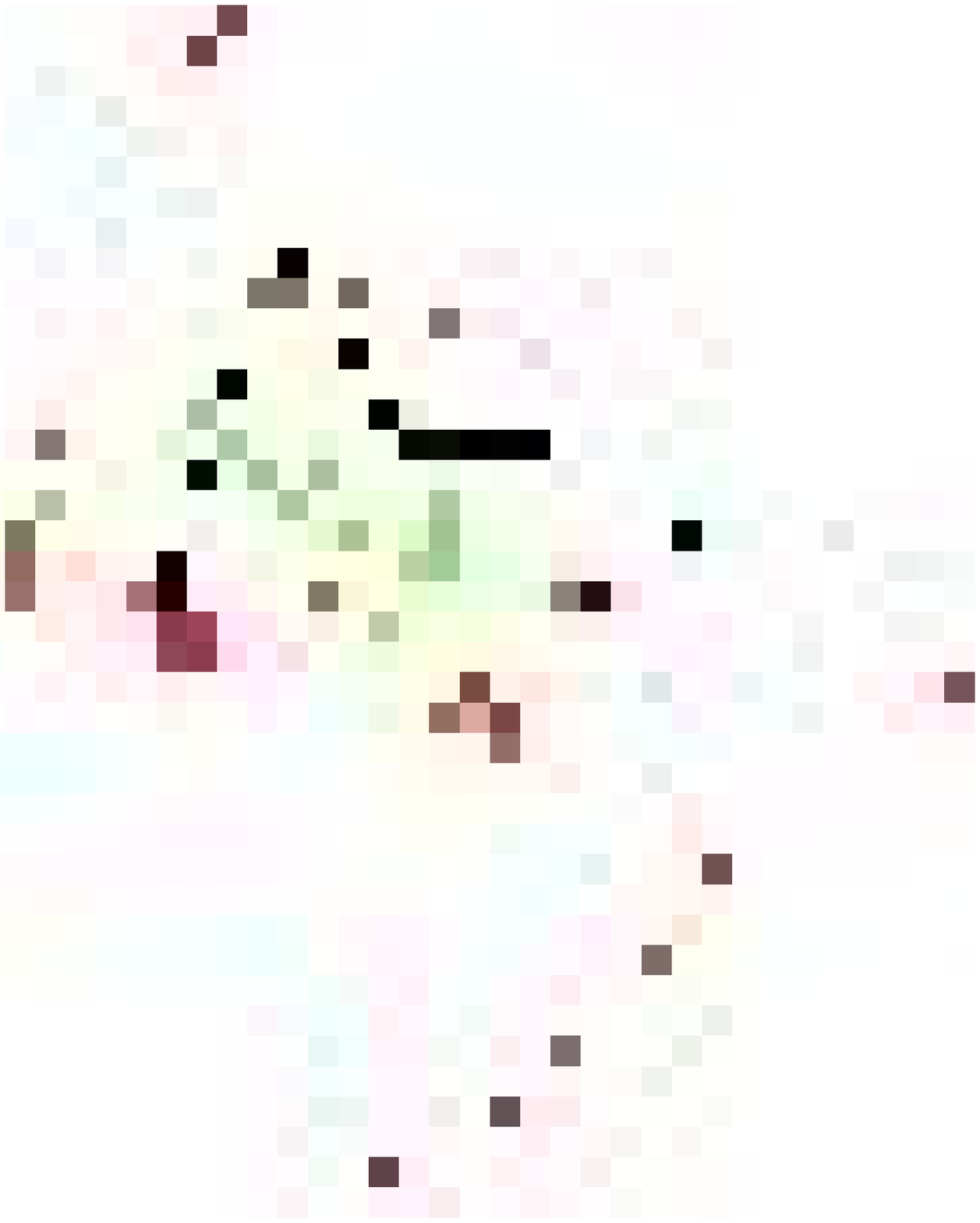}}}
           \caption{Aspects 4 and 5}
           \label{figure:Aspect_4_5}
       \end{minipage} \\ \\
       \begin{minipage}[t]{40 mm}
           \centerline{\hbox{\includegraphics[width= 20mm,height= 20mm]{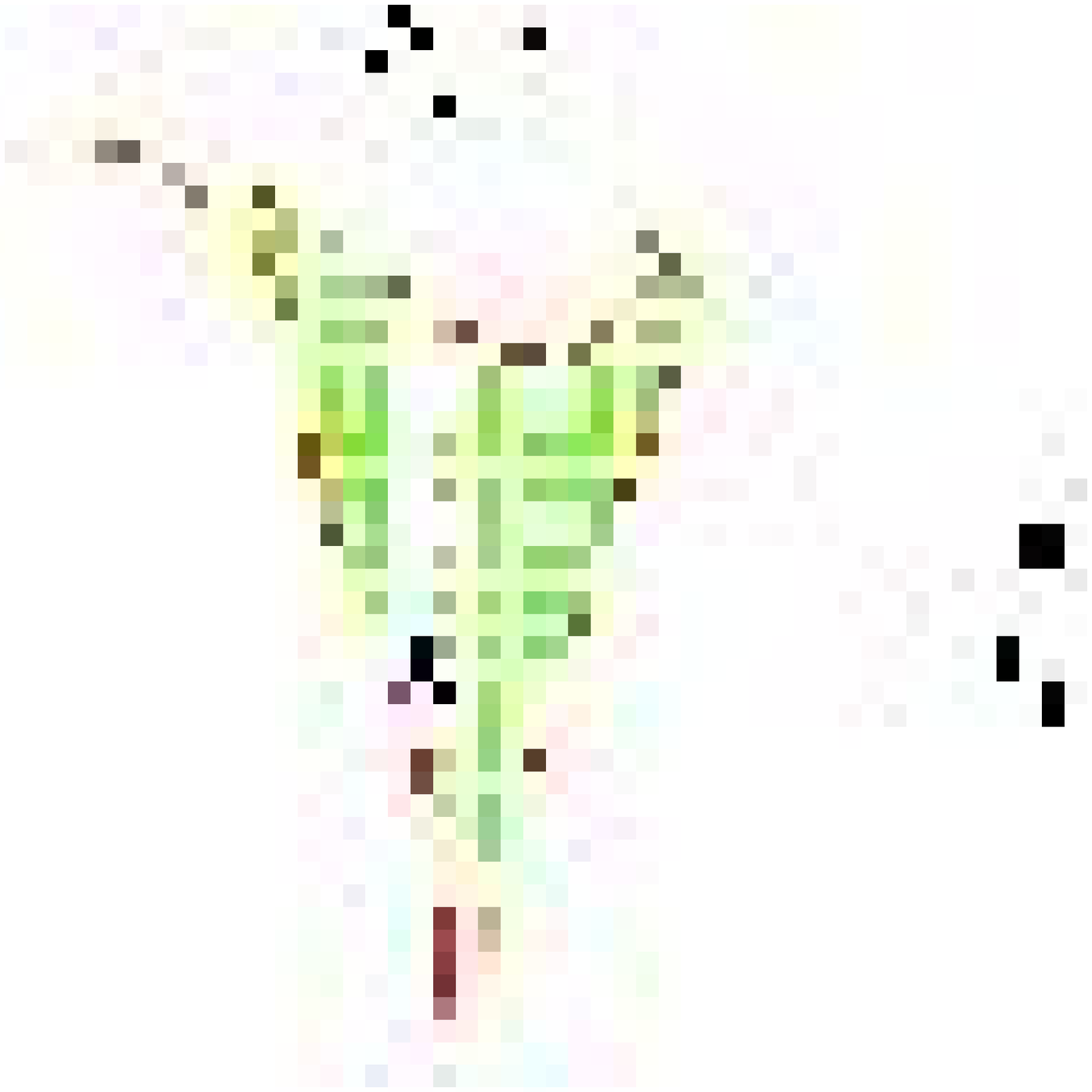}
                             \includegraphics[width= 20mm,height= 20mm]{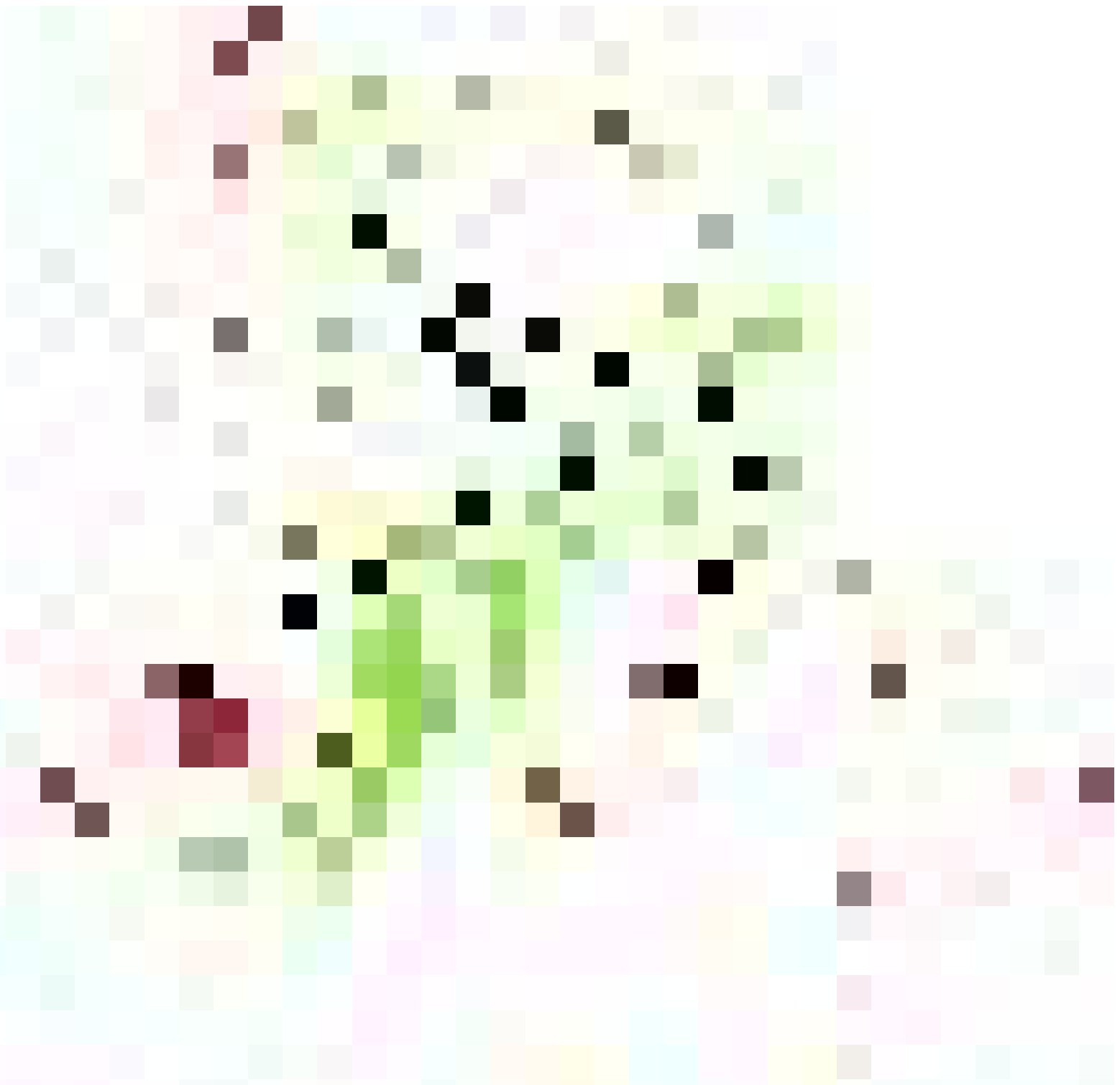}}}
           \caption{Aspect 6}
           \label{figure:Aspect_6}
       \end{minipage} &
       \begin{minipage}[t]{42 mm}
           \centerline{\hbox{\includegraphics[width= 20mm,height= 20mm]{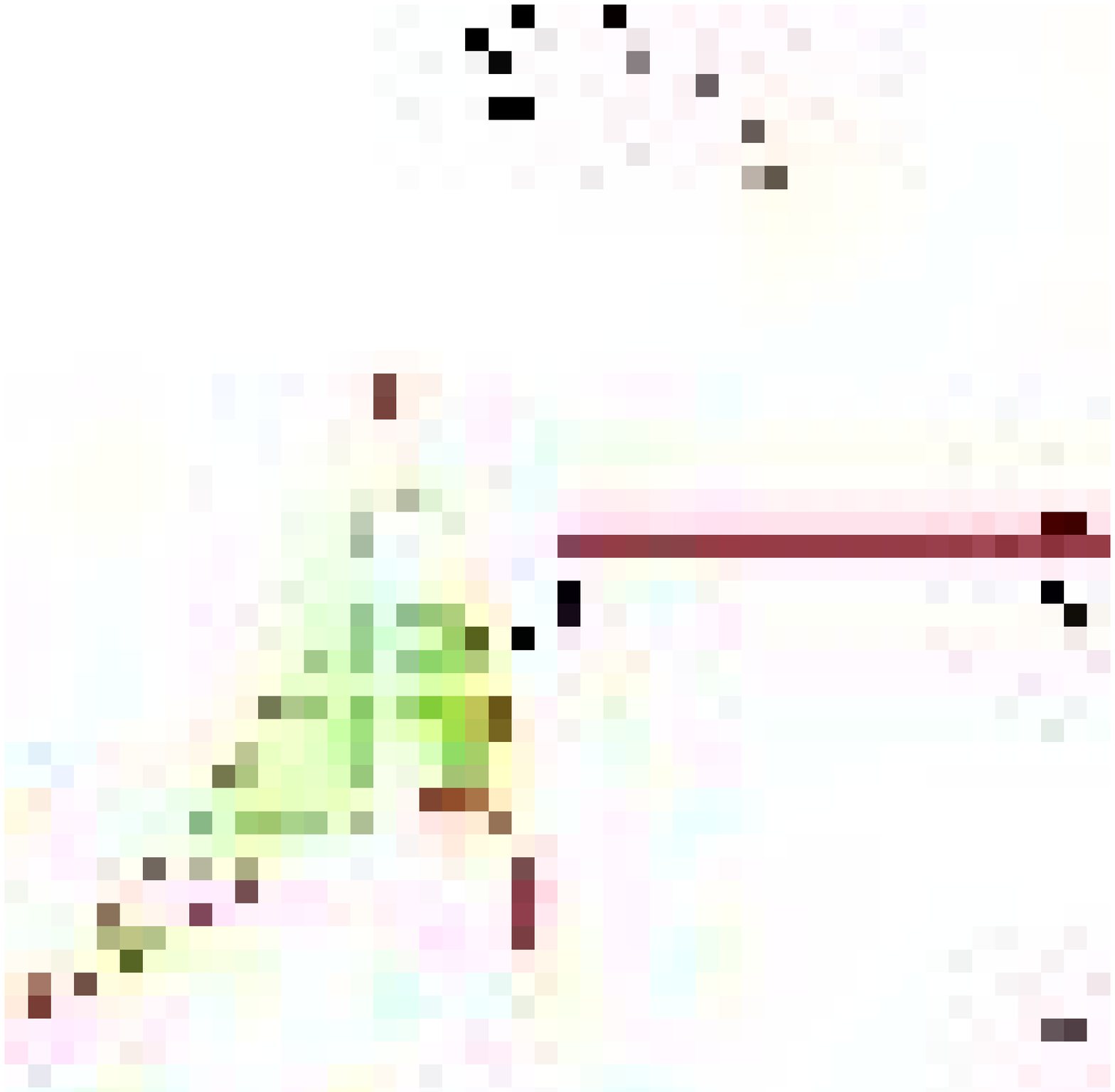}
                             \includegraphics[width= 20mm,height= 20mm]{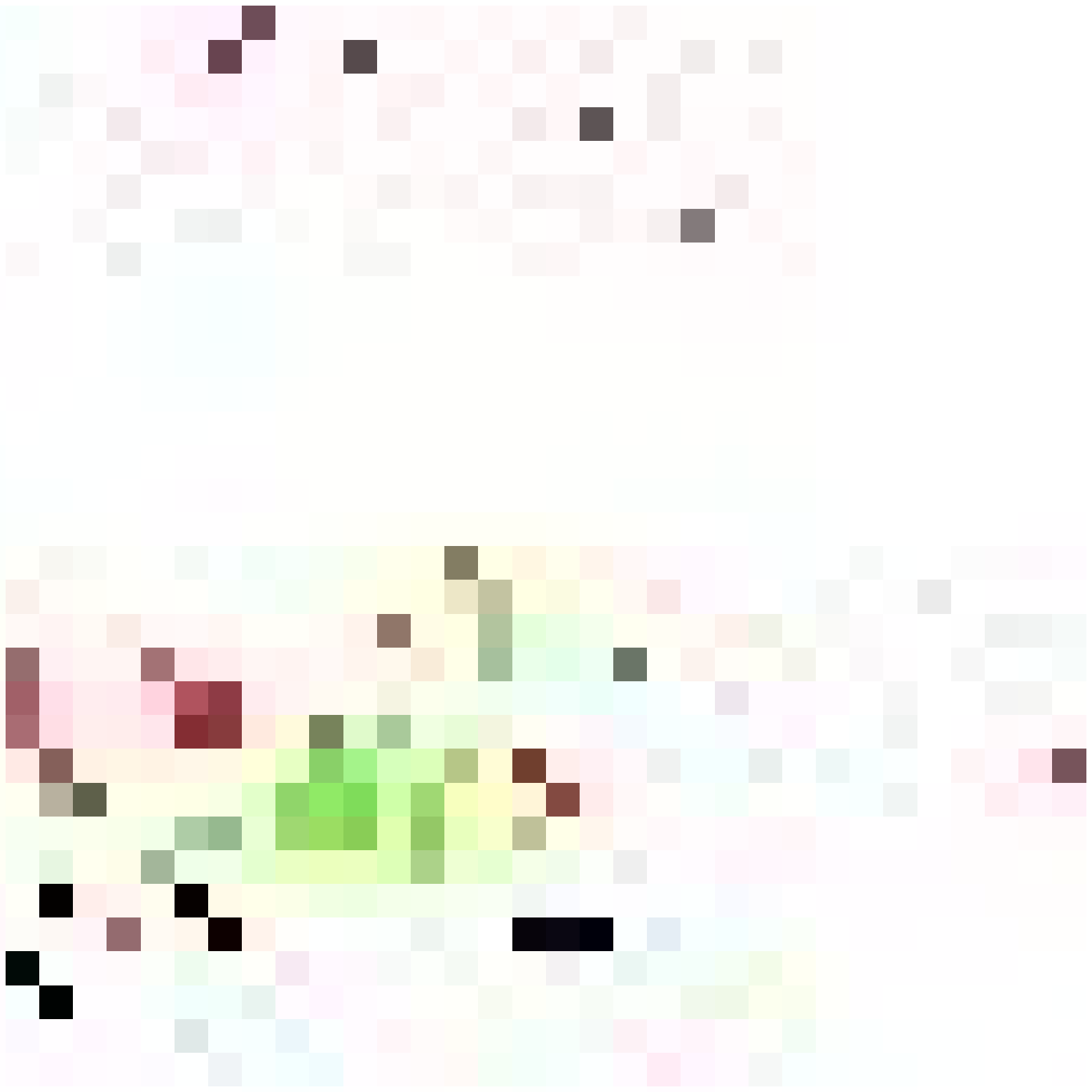}}}
           \caption{Aspects 7 and 8}
           \label{figure:Aspect_7_8}
       \end{minipage} \\ \\
       \begin{minipage}[t]{40 mm}
           \centerline{\hbox{\includegraphics[width= 20mm,height= 20mm]{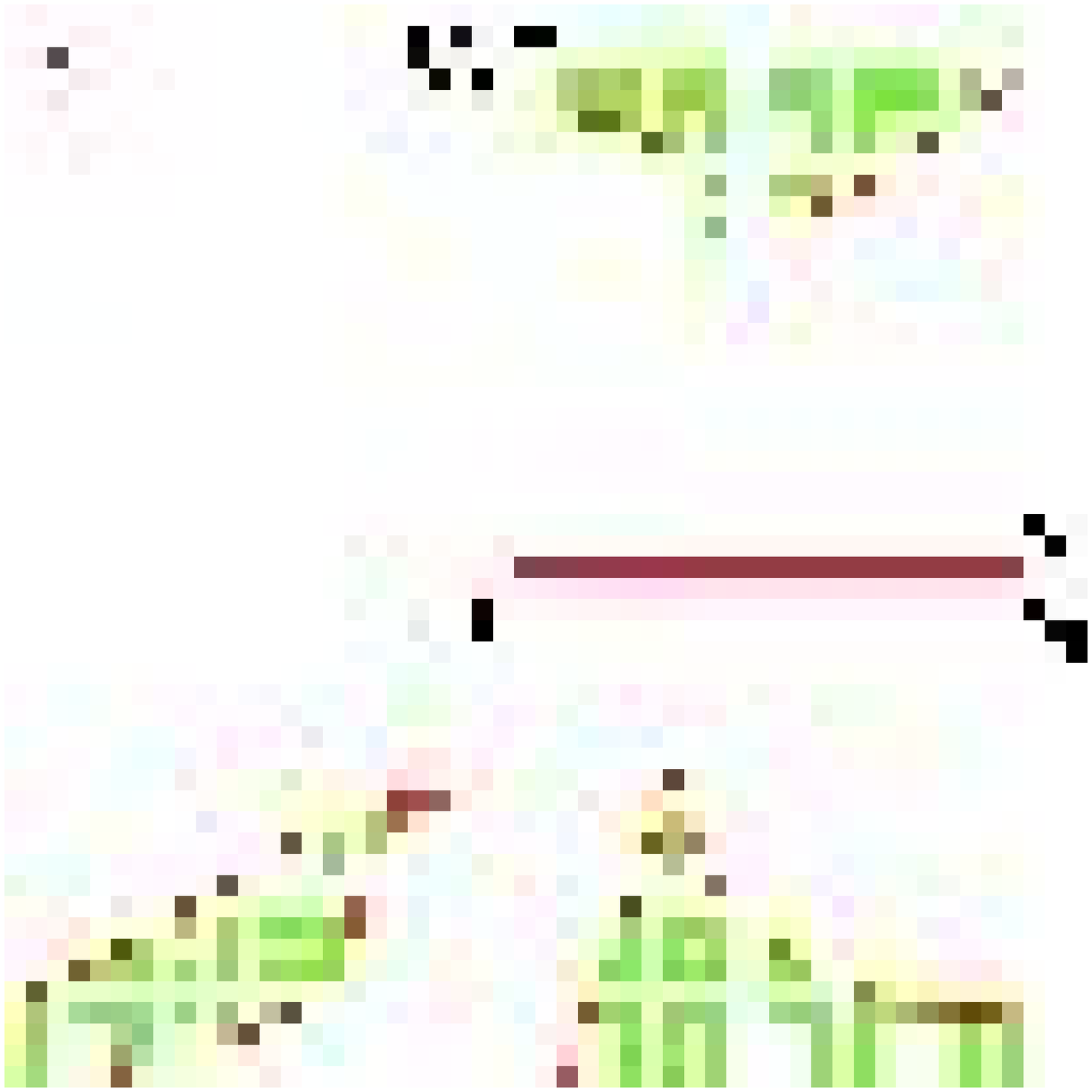}
                             \includegraphics[width= 20mm,height= 20mm]{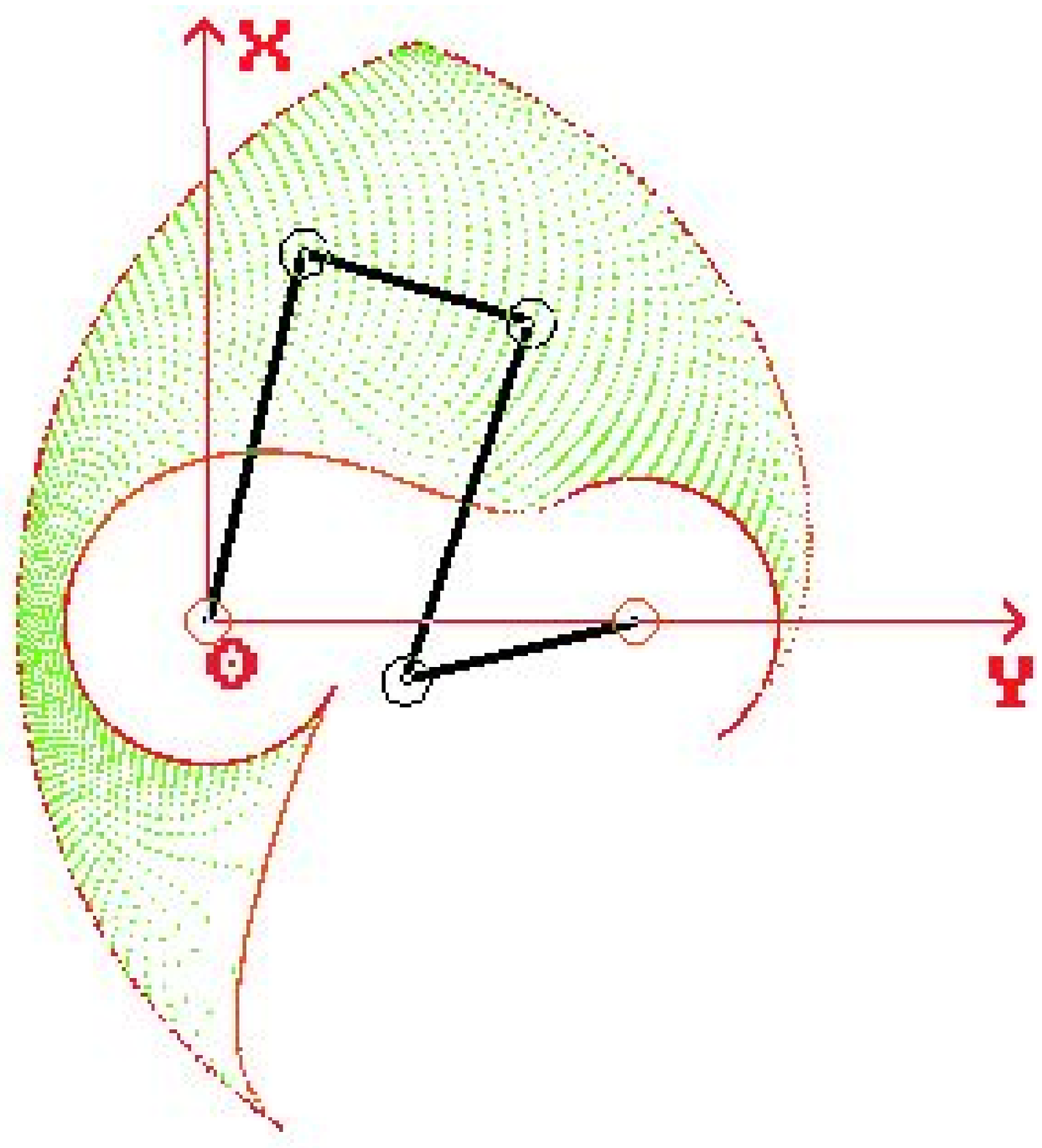}}}
           \caption{Aspect 9}
           \label{figure:Aspect_9}
       \end{minipage} &
       \begin{minipage}[t]{42 mm}
           \centerline{\hbox{\includegraphics[width= 20mm,height= 20mm]{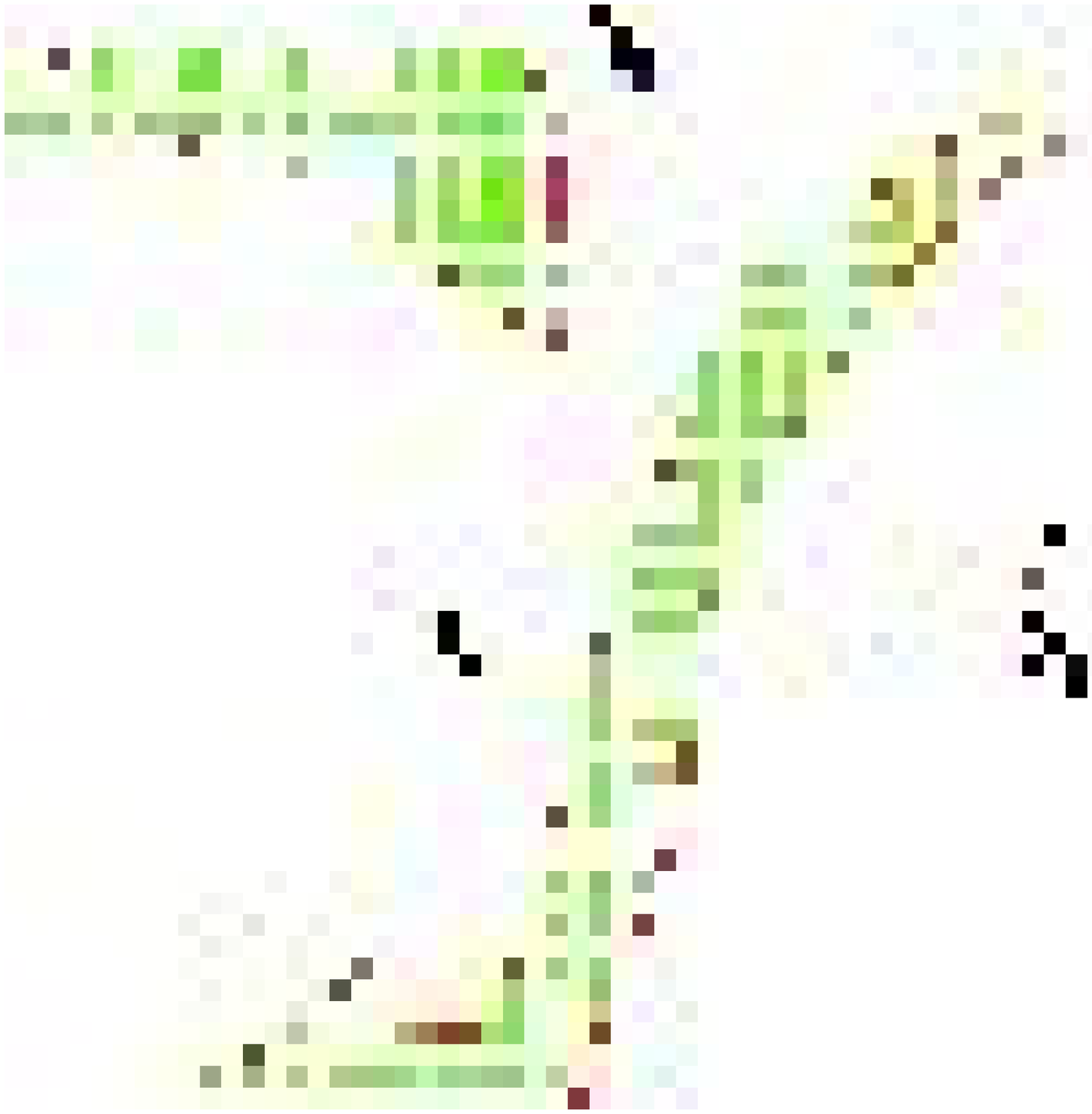}
                             \includegraphics[width= 20mm,height= 20mm]{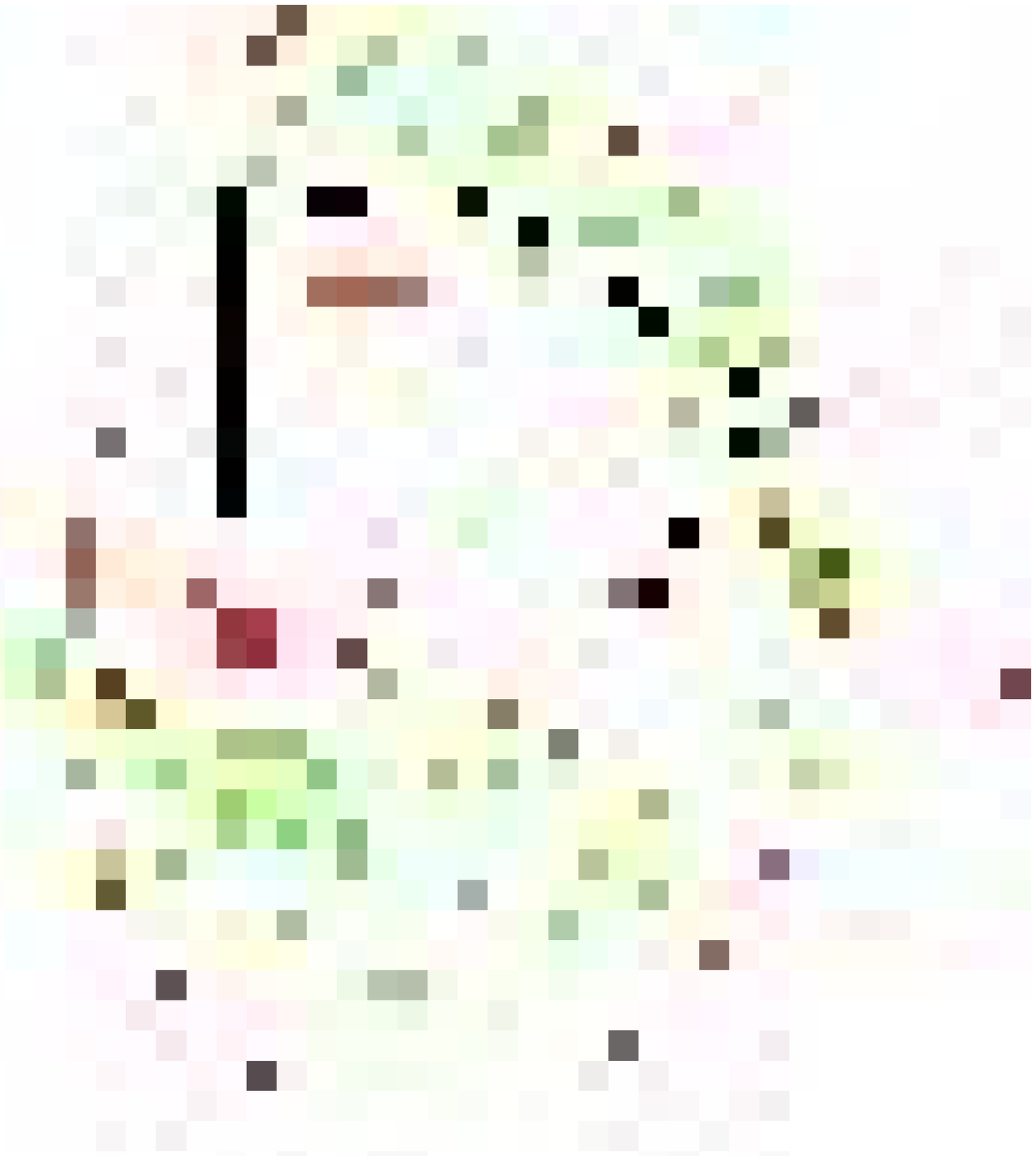}}}
           \caption{Aspect 10}
           \label{figure:Aspect_10}
       \end{minipage}
    \end{tabular}
    \end{center}
\end{figure}

For the manipulator at hand, the generalized aspects are defined
with the definition~\ref{definition:Aspect}. Figures
\ref{figure:Aspect_1}-\ref{figure:Aspect_10} depict the different
serial and parallel aspects obtained. Table \ref{table:aspect}
shows that there are 10 serial/parallel aspects for the manipulator
(N and P stand for negative and positif, respectively).
\par
In opposite to the aspects defined by \cite{Borrel:86} or by
\cite{Wenger:97}, the generalized aspects are not disjoint. Only,
the aspects belonging to the same working mode are disjoint. In the
sample in which the manipulator has only one inverse kinematic
solution, the notion of generalized aspect is equivalent to the
notion of aspect given by \cite{Wenger:97}. Indeed, if the
manipulator has only one working mode (like the 2-RPR planar
manipulator), then the parallel aspects are disjoint and represent
the maximal singularity-free domain in the Cartesian workspace.
\begin{table}[hbt]
\begin{tabular}{|c|c|c|c|c|c|c|c|c|} \hline
$Figures$      & \ref{figure:Aspect_1} & \ref{figure:Aspect_2}   &
                 \ref{figure:Aspect_3} & \ref{figure:Aspect_4_5} &
                 \ref{figure:Aspect_6} & \ref{figure:Aspect_7_8} &
                 \ref{figure:Aspect_9} & \ref{figure:Aspect_10}  \\ \hline
$det(\negr A)$ & P & P & P & P & N & N & N & N  \\ \hline

$\negr B_{11}$ & P & P & N & N & P & P & N & N  \\ \hline

$\negr B_{22}$ & P & N & N & P & N & P & P & N  \\ \hline
$\begin{array}{c}
   Nb~of\\
\!\!\!generalized\!\!\!\\
   aspects
\end{array}$   & 1 & 1 & 1 & 2 & 1 & 2 & 1 & 1 \\ \hline
\end{tabular}
\caption{The generalized aspects}
\label{table:aspect}
\end{table}
\section{Conclusions}
In this paper, the notion of aspect was defined for parallel
manipulators with multiple inverse and direct kinematic solutions.
The working modes were introduced to define this notion. For one
working mode, we can find out the maximal singularity-free domains
of the Cartesian product of the workspace with the joint space.
This work brings material to further investigations like trajectory
planning and kinematic design which are the subject of current
research work from the authors.
\par
The generalized aspect are not the uniqueness domain in any case as
it was shown in \cite{Wenger:97}. So, in a future study, the
authors will define uniqueness domains for general fully parallel
manipulators. In such domains, there are only one inverse and
direct kinematic solution. These domains are of interest for the
control of manipulator.
\goodbreak
\bibliographystyle{unsrt}

\end{document}